\documentclass{article}

\PassOptionsToPackage{numbers, compress}{natbib}

\usepackage[preprint]{neurips_2025}

\usepackage[utf8]{inputenc} 
\usepackage[T1]{fontenc}    
\usepackage{hyperref}       
\usepackage{url}            
\usepackage{booktabs}       
\usepackage{amsfonts}       
\usepackage{nicefrac}       
\usepackage{microtype}      
\usepackage{xcolor}         

\usepackage{amsmath,amsfonts,bm}

\def\1{\bm{1}}

\def\va{{\bm{a}}}

\def\vr{{\bm{r}}}
\def\vs{{\bm{s}}}

\def\vu{{\bm{u}}}

\def\vw{{\bm{w}}}

\DeclareMathAlphabet{\mathsfit}{\encodingdefault}{\sfdefault}{m}{sl}
\SetMathAlphabet{\mathsfit}{bold}{\encodingdefault}{\sfdefault}{bx}{n}

\DeclareMathOperator*{\argmax}{arg\,max}

\usepackage{url}
\usepackage{multirow}
\usepackage{diagbox}
\usepackage{amsmath,graphicx,amssymb,mathtools,amsthm,subfigure,float,algorithm,algorithmic,mathtools,dsfont}

\definecolor{mydarkblue}{rgb}{0,0.08,0.45}
\definecolor{mydarkgreen}{RGB}{0, 139, 69}
\hypersetup{
	colorlinks=true,
	urlcolor=magenta,
	citecolor=mydarkblue,
}

\theoremstyle{plain}
\newtheorem{theorem}{Theorem}[section]

\theoremstyle{definition}

\theoremstyle{remark}

\title{Exploratory Diffusion Model for \\ Unsupervised Reinforcement Learning}

\author{%
   Chengyang Ying \quad Huayu Chen \quad Zhongkai Hao \quad Xinning Zhou \\
   \textbf{Hang Su} \quad \textbf{Jun Zhu}\\
   Department of Computer Science \& Technology, Institute for AI, BNRist Center,\\
   Tsinghua-Bosch Joint ML Center, THBI Lab, Tsinghua University;\\
}

\begin{document}

\maketitle

\begin{abstract}
Unsupervised reinforcement learning (URL) aims to pre-train agents by exploring diverse states or skills in reward-free environments, facilitating efficient adaptation to downstream tasks. 
As the agent cannot access extrinsic rewards during unsupervised exploration, existing methods design intrinsic rewards to model the explored data and encourage further exploration.
However, the explored data are always heterogeneous, posing the requirements of powerful representation abilities for both intrinsic reward models and pre-trained policies.
In this work, we propose the Exploratory Diffusion Model (\textbf{ExDM}), which leverages the strong expressive ability of diffusion models to fit the explored data, simultaneously boosting exploration and providing an efficient initialization for downstream tasks.
Specifically, ExDM can accurately estimate the distribution of collected data in the replay buffer with the diffusion model and introduces the score-based intrinsic reward, encouraging the agent to explore less-visited states.
After obtaining the pre-trained policies, ExDM enables rapid adaptation to downstream tasks.
In detail, we provide theoretical analyses and practical algorithms for fine-tuning diffusion policies, addressing key challenges such as training instability and computational complexity caused by multi-step sampling.
Extensive experiments demonstrate that ExDM outperforms existing SOTA baselines in efficient unsupervised exploration and fast fine-tuning downstream tasks, especially in structurally complicated environments.
\end{abstract}

\section{Introduction}
\label{sec_intro}


Developing generalizable agents of efficient adaptation across diverse tasks remains a fundamental challenge in reinforcement learning (RL). 
To address this, unsupervised RL (URL)~\cite{eysenbach2018diversity,laskin2021urlb} pre-trains agents in reward-free environments to acquire diverse skills or general representations that facilitate fast adaptation to downstream tasks with limited interaction.
Unlike standard RL, which benefits from task-specific rewards, URL relies on carefully designed intrinsic rewards to guide exploration in reward-free environments. 
This makes URL especially dependent on accurate modeling of complex, heterogeneous data distributions, demanding greater expressiveness and generalizability of agents.


One of the major challenges in URL is the requirement of strong modeling and fitting abilities for both pre-training and fine-tuning. 
During pre-training, maximizing exploration in reward-free environments requires intrinsic rewards based on an accurate estimation of the collected data distribution. 
However, this distribution is often heterogeneous, highlighting the importance of modeling ability.
While existing URL methods can collect diverse trajectories, they typically employ simple pre-trained policies, such as Gaussian policies~\cite{pathak2017curiosity,mazzaglia2022curiosity} or discrete skill-based policies~\cite{eysenbach2018diversity,laskin2022unsupervised}, due to their ease of training and sampling.
Consequently, these methods often fail to reflect the full diversity of the explored data in the replay buffer, limiting effective unsupervised exploration and downstream fine-tuning. 
Increasing model expressiveness, like deep generative models, could mitigate this limitation, but may introduce additional challenges like unstable training and high computing costs. 
Overcoming these issues remains critical for advancing the capabilities and practicality of URL.

To address the above challenges, we propose \textbf{Ex}ploratory \textbf{D}iffusion \textbf{M}odel (\textbf{ExDM}), which leverages the powerful modeling capabilities of diffusion models~\cite{song2021scorebased,chi2023diffusion} while maintaining training efficiency, to enhance unsupervised exploration and fast adaptation. 
During unsupervised pre-training, ExDM employs a diffusion model to accurately model the diverse and always heterogeneous state distribution in the replay buffer, overcoming the challenge typically encountered in URL. 
Based on this accurate distribution modeling, we propose a score-based intrinsic reward $\mathcal{R}_{\mathrm{score}}$ that explicitly guides the agent to explore less-visited regions, thus maximizing the state distribution entropy and promoting more effective exploration.
However, directly sampling actions from diffusion models may cause inefficient computing costs due to the multi-step sampling.
To address this, ExDM employs an additional Gaussian behavior policy for efficient action selection. This behavior policy is trained to maximize the score-based intrinsic reward, ensuring efficient exploration without incurring significant computational overhead.


In addition to enhancing unsupervised exploration, ExDM further provides powerful policy initialization for efficient fine-tuning of downstream tasks.
Besides fine-tuning the Gaussian behavior policy by standard RL algorithms, ExDM can also adapt the pre-trained diffusion policies to downstream tasks.
The fine-tuning stage in URL requires only allows limited steps of online interaction, posing extra challenges for adapting diffusion policies.
To address this, we theoretically analyze the fine-tuning objective, resulting in an alternating method, of which the optimality is formally established in Theorem~\ref{thm_2}.

\begin{figure*}[t]
\centering
\includegraphics[width=\linewidth]
{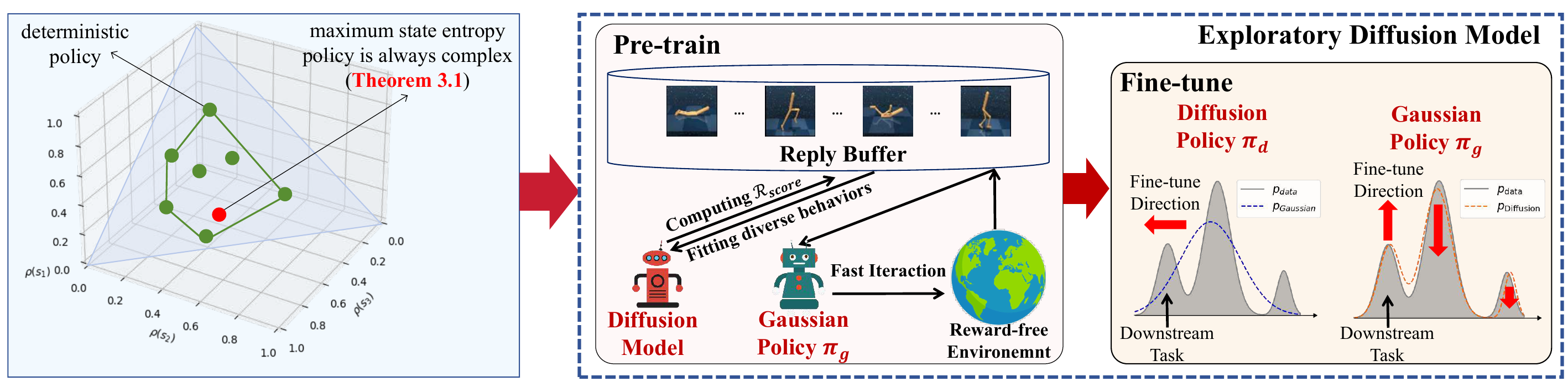}
\vspace{-1.em}
\caption{
\textbf{Overview of Exploratory Diffusion Model (ExDM).}
Different from standard RL, URL aims to explore in reward-free environments, requiring expressive policies and models to fit heterogeneous data (Theorem~\ref{thm_1}).
During pre-training, ExDM employs the diffusion model to model the heterogeneous exploration data and calculate score-based intrinsic rewards to encourage exploration.
Moreover, we adopt a Gaussian behavior policy to collect data that avoids the inefficiency caused by the multi-step sampling of the diffusion policy. 
}
\label{fig_overview}
\vspace{-1.em}
\end{figure*}

We evaluate ExDM's performance in both unsupervised exploration and downstream task adaptation across various benchmarks, including Maze2d~\cite{campos2020explore} and continuous control in URLB~\cite{laskin2022unsupervised}. 
Both qualitative visualization and quantitative metric in Maze2d demonstrate that ExDM achieves a substantially larger state coverage ratio during pre-training than existing baselines, highlighting our superior exploration capability.
Notably, in the most complicated mazes, compared to the SOTA baselines, ExDM can improve performance by 51$\%$ and achieve comparable performance only with 37$\%$ timesteps.
These challenging mazes, featuring numerous branching paths and decision points, reveal ExDM's ability to efficiently explore diverse regions within limited timesteps, whereas baselines often struggle at some wall corners and fail to cover the entire maze.
Moreover, extensive experiments in URLB show that ExDM can efficiently adapt to downstream tasks and outperform URL baselines, as well as diffusion policy fine-tuning baselines, by large margins.

In summary, the main contributions are as follows:
\begin{itemize}
    \item To the best of our knowledge, this is the first work to leverage diffusion models in URL, efficiently capturing heterogeneous data distributions and boosting unsupervised exploration. 
    \item Besides exploration, ExDM provides theoretical analyses and practical algorithms to fast adapt pre-trained diffusion policies to downstream tasks with limited interactions. 
    \item Extensive evaluation across various domains demonstrates that ExDM can achieve efficient exploration and fast fine-tuning, obtaining SOTA in various settings.
\end{itemize}

\section{Background}

\subsection{Unsupervised Reinforcement Learning}

Reinforcement learning (RL) considers Markov decision processes (MDP) $\mathcal{M} = (\mathcal{S},\mathcal{A}, \mathcal{P}, \mathcal{R}, \rho_0, \gamma)$ \cite{sutton2018reinforcement}. Here $\mathcal{S}$ and $\mathcal{A}$ denote the state and action spaces, respectively. For $\forall (\vs,\va)\in\mathcal{S}\times \mathcal{A}$, $\mathcal{P}(\cdot|\vs,\va)$ is a distribution on $\mathcal{S}$, representing the dynamic of $\mathcal{M}$, and $\mathcal{R}(\vs, \va)$ is the extrinsic task reward function. $\rho_0$ is the initial state distribution and $\gamma$ is the discount factor. For a given policy $\pi: \mathcal{S}\rightarrow \Delta(\mathcal{A})$, we define the discount state distribution of $\pi$ at state $\vs$ as $d_{\pi}(\vs) = (1-\gamma)\sum_{t=0}^{\infty} \left[ \gamma^t \mathcal{P}(\vs^t=\vs)\right]$.
The objective of RL is to maximize the expected cumulative return of $\pi$ over the task $\mathcal{R}$:
\begin{equation}
    J(\pi) \triangleq \mathbb{E}_{\tau\sim \mathcal{M},\pi} \left[\mathcal{R}(\tau)\right] = \frac{1}{1-\gamma}\mathbb{E}_{\vs\sim d_{\pi}, \va\sim\pi} \left[\mathcal{R}(\vs, \va)\right].
\end{equation}
To boost agents' generalization, unsupervised RL (URL) typically includes two stages: \emph{unsupervised pre-training} and \emph{few-shot fine-tuning}. 
During pre-training, agents explore the reward-free environment $\mathcal{M}^c$, i.e., $\mathcal{M}$ without the reward function $\mathcal{R}$. 
Thus, URL requires designing intrinsic rewards $\mathcal{R}_{\mathrm{int}}$ to guide policies to maximize the state entropy $\mathcal{H}(d_{\pi}(\cdot))$.
During fine-tuning, agents adapt pre-trained policies to handle the downstream task represented by the extrinsic task-specific reward $\mathcal{R}$, through limited online interactions (like one-tenth or less of pre-training steps, the formulation is in Eq.~\ref{eq_finetune_obj}).

\subsection{Diffusion Models in Reinforcement Learning}

Recent studies have demonstrated that diffusion models~\cite{sohl2015deep,ho2020denoising} excel at accurately representing heterogeneous behaviors in continuous control, particularly through the use of diffusion policies~\cite{chen2023offline,wang2023diffusion,chi2023diffusion}. Given state-action pairs $(\vs, \va)$ sampled from some unknown policy $\mu(\va|\vs)$, diffusion policies consider the forward diffusion process that gradually injects standard Gaussian noise $\bm{\epsilon}$ into actions:
\begin{equation}
    \va_t = \alpha_t \va + \sigma_t \bm{\epsilon},\quad t\in [0,1],
\end{equation}
here $\alpha_t, \sigma_t$ are pre-defined hyperparameters satisfying that when $t=0$, we have $\va_t = \va$, and when $t=1$, we have $\va_t \approx \bm{\epsilon}$.
For $\forall t \in [0,1]$, we can define the marginal distribution of $\va_t$ as
\begin{equation}
    p_t (\va_t | \vs, t) = \int \mathcal{N}(\va_t|\alpha_t \va, \sigma_t^2 \bm{I}) \mu(\va | \vs) d \va.
\end{equation}
Then we train a conditional ``noise predictor" $\bm{\epsilon}_{\theta}(\va_t | \vs, t)$ to predict the added noise of each timestep:
\begin{equation}
\label{eq_diff_policy}
    \min_{\theta} \mathbb{E}_{t, \bm{\epsilon}, \vs, \va} [\|\bm{\epsilon}_{\theta} (\va_t | \vs, t) - \bm{\epsilon}\|^2].
\end{equation}
After training $\bm{\epsilon}_\theta$, we can discretize the diffusion ODEs of the reverse process~\cite{song2021scorebased} and sample actions with numerical solvers~\cite{song2021denoising,lu2022dpm} in around $5\sim 15$ steps, to approximate the original policy $\mu(\va|\vs)$.
However, this multi-step sampling affects the training efficiency, especially in online settings.

\section{Methodology}
\label{sec_method}
Below, we introduce the Exploratory Diffusion Model (ExDM) to capture heterogeneous data to boost unsupervised exploration (Sec.~\ref{sec_pretrain}) and obtain powerful initialization for fast fine-tuning (Sec.~\ref{sec_finetune}). 


\subsection{Exploratory Diffusion Model for Unsupervised Pre-training}
\label{sec_pretrain}


The major challenge and objective during unsupervised pre-training is to explore diverse states in reward-free environments. Consequently, a natural pathway is to pre-train the policy to maximize the entropy of the state~\cite{liu2021behavior,lu2023contrastive}, i.e., $\mathcal{H}(d_{\pi}(\cdot)) = \int_{s} -d_{\pi}(s) \log d_{\pi}(s) d s$.
Although it is well known that the optimal policy of standard RL is a simple deterministic policy, we first prove that, even if the environment is discrete, the policies with the maximum state entropy are still complicated and not deterministic with a high probability, requiring much stronger modeling abilities. 
\begin{theorem}[Details and proof are in Appendix~\ref{app_proof_thm_1}]
\label{thm_1}
    When $\mathcal{S}, \mathcal{A}$ are discrete spaces, i.e., $|\mathcal{S}| = S, |\mathcal{A}| = A$, there are $M\triangleq A^S$ deterministic policies. Set $\hat{\pi} = \argmax_{\pi} \mathcal{H}(d_{\pi}(\cdot))$, under some mild assumptions, we have 
    \begin{equation}
    \begin{split}
        P(\hat{\pi}\text{ is not deterministic policy and }\mathcal{H}(d_{\hat{\pi}})=\log |S|) \ge & 1 - M^{S} v(S)^M,
    \end{split}
    \end{equation}
    will fast converge to 1 with the increasing of $A$, and here $v(S)$ is a constant only related to $S$ and satisfies $0 < v(S) < 1$.
\end{theorem}



\begin{algorithm}[t]
    \caption{Pre-training of ExDM} 
    \label{algo_ExDM_pretrain}
    \begin{algorithmic}[1] 
        \REQUIRE Reward-free environment $\mathcal{M}^c$, replay buffer $\mathcal{D}$, Gaussian behavior policy $\pi_{\mathrm{g}}$, diffusion policy $\pi_{\mathrm{d}}$ parameterized with the score model $\bm{\epsilon}_{\theta}$, state diffusion model $\bm{\epsilon}_{\theta'}$.
        \OUTPUT Pre-trained Gaussian behavior policy $\pi_{\mathrm{g}}$ and pre-trained diffusion policy $\pi_{\mathrm{d}}$
        \FOR{$\text{sample step} = 1,2,..., S$}
        \FOR{$\text{update step} = 1,2,..., U$}
        \STATE Sample $\vs$-$\va$ pairs $\{(\vs^m,\va^m)\}_{m=1}^{M}$ from $\mathcal{D}$.
        \STATE Update $\bm{\epsilon}_{\theta}$ and $\bm{\epsilon}_{\theta'}$ via optimizing with Eq.~(\ref{eq_train_state_score}) with sampled data.
        \STATE Calculate the score-based intrinsic rewards $\vr^m$ via Eq.~(\ref{eq_score_rew}) for each sampled pair $(\vs^m, \va^m)$.
        \STATE Train $\pi_{\mathrm{g}}$ with $(\vs^m, \va^m, \vr^m)$ by any RL algorithm.
        \ENDFOR
        \STATE Utilize the behavior policy $\pi_{\mathrm{g}}$ to interact with $\mathcal{M}^c$ and store state-action pairs into $\mathcal{D}$.
        \ENDFOR
    \end{algorithmic}  
\end{algorithm} 

The above theorem demonstrates that maximizing state entropy requires policies with strong expression abilities, rather than simple deterministic policies.
Moreover, despite previous work mainly considering simple Gaussian policies or skill-based policies, the explored replay buffer is always diverse and heterogeneous, as the policy continuously changes to visit new states during pre-training.
Consequently, during unsupervised pre-training, URL requires capturing the heterogeneous distribution of collected data and obtaining policies with high diversity.
These challenges pose the requirement of strong density estimation and fitting abilities, while maintaining training stability and efficiency.
Inspired by the recent great success of diffusion models in modeling diverse image distributions~\cite{dhariwal2021diffusion,saharia2022photorealistic} and behaviors~\cite{chi2023diffusion,janner2022planning}, ExDM proposes to utilize the diffusion models $\bm{\epsilon}_{\theta'}$ and $\bm{\epsilon}_{\theta}$ to model the distribution of states and state-action pairs in the replay buffer $\mathcal{D}$ collected before:
\begin{equation}
\begin{split}
\label{eq_train_state_score}
    & \min \mathbb{E}_{\vs, \va \sim \mathcal{D}} [\mathbb{E}_{t,\bm{\epsilon}} \|\bm{\epsilon}_{\theta'} (\vs_t | t) - \bm{\epsilon}\|^2 +  \mathbb{E}_{t,\bm{\epsilon}} \|\bm{\epsilon}_{\theta} (\va_t | \vs, t) - \bm{\epsilon}\|^2].
\end{split}
\end{equation}
To maximize the entropy of the state distribution, we can use $\log p_{\theta'}(\vs)$ to measure the frequency of states in the replay buffer. Consequently, we design $-\log p_{\theta'}(\vs)$ as the intrinsic reward to encourage the agent to explore these regions. 
Although estimating the log-probability of the diffusion model is challenging, it is well known that $-\log p_{\theta'}(\vs)$ can be bounded by the following evidence lower bound (ELBO)~\cite{ho2020denoising}:
\begin{equation}
\begin{split}
    -\log p_{\theta'}(\vs) \leq \mathbb{E}_{\bm{\epsilon}, t} [\vw_t\|\bm{\epsilon}_{\theta'}(\vs_t| t) - \bm{\epsilon}\|^2] + C,
\end{split}
\end{equation}
here $C$ is a constant independent of $\theta'$, and $\vw_t$ are parameters related to $\alpha_t,\sigma_t$, which are typically ignored~\cite{ho2020denoising}. Consequently, we propose our score-based intrinsic rewards as:
\begin{equation}
\begin{split}
\label{eq_score_rew}
    & \mathcal{R}_{\mathrm{score}}(\vs) = \mathbb{E}_{\bm{\epsilon}, t} [\| \bm{\epsilon}_{\theta'}(\vs | t) - \bm{\epsilon}\|^2].
\end{split}
\end{equation}
Intuitively, our score-based intrinsic rewards can measure the fitting quality of the diffusion model to the explored data, thereby encouraging the agent to explore regions that are poorly fitted or unexplored.
By maximizing these intrinsic rewards, ExDM trains agents to discover unseen regions effectively.
However, directly using diffusion policies to interact with reward-free environments during pre-training is inefficient and unstable due to the requirement of multi-step sampling.
To address this limitation, ExDM incorporates a Gaussian behavior policy $\pi_{\mathrm{g}}$ for efficient action sampling.
The Gaussian behavior policy  $\pi_{\mathrm{g}}$ can then be trained using any RL algorithm, guided by the score-based intrinsic reward $\mathcal{R}_{\mathrm{score}}(\vs)$.
This encourages the exploration of regions where the diffusion model either fits poorly or has not yet been exposed.
The pseudo code of the unsupervised exploration stage of ExDM is in Algorithm~\ref{algo_ExDM_pretrain}.




\subsection{Efficient Online Fine-tuning to Downstream Tasks}
\label{sec_finetune}

When adapting pre-trained policies to downstream tasks with limited timesteps, existing URL methods always directly apply online RL algorithms like DDPG~\cite{lillicrap2015continuous} or PPO~\cite{schulman2017proximal} for fine-tuning.
The behavior policy $\pi_{\mathrm{g}}$ in ExDM can also be fine-tuned to handle the downstream task with the same online RL algorithms, performing fair comparison of exploration efficiency between ExDM and baselines (detailed experimental results are in Sec.~\ref{sec_exp_finetune_gaussian}).

Moreover, besides the behavior policy $\pi_{\mathrm{g}}$, ExDM has also pre-trained the diffusion policy $\pi_{\mathrm{d}}$, which can better capture the heterogeneous explored trajectories and can adapt to downstream tasks. 
Unfortunately, it is challenging to online fine-tune the diffusion policy due to the instability caused by the multi-step sampling and the lack of closed-form probability calculation~\cite{chen2024aligning,ren2024diffusion}.
To address these challenges, we first analyze the online fine-tuning objective for URL.
Given the limited iteration timesteps during fine-tuning, the objective can be formulated as the combination of maximizing the cumulative return and keeping close to the pre-trained policy over all $\vs$~\cite{eysenbach2021information,ying2024peac}:
\begin{equation}
\begin{split}
\label{eq_finetune_obj}
    \max_{\pi} J_{\mathrm{f}}(\pi) \triangleq  & J(\pi) - \frac{\beta}{(1-\gamma)} \mathbb{E}_{\vs\sim d_{\pi}}  \left[D_{\mathrm{KL}}(\pi(\cdot|\vs) \| \pi_{\mathrm{d}}(\cdot|\vs) ) \right] \\
    = & \frac{1}{1-\gamma} \mathbb{E}_{\vs\sim d_{\pi} , \va \sim \pi}  \left[\mathcal{R}(\vs, \va) -  \beta D_{\mathrm{KL}}(\pi(\cdot|\vs) \| \pi_{\mathrm{d}}(\cdot|\vs) ) \right] \\
    = & \frac{1}{1-\gamma} \mathbb{E}_{\vs\sim d_{\pi} , \va \sim \pi}  \left[\mathcal{R}(\vs, \va) -  \beta \log\frac{\pi(\va|\vs)}{\pi_{\mathrm{d}}(\va|\vs)} \right],
\end{split}
\end{equation}
here $\beta > 0$ is an unknown trade-off parameter that is related to the fine-tuning steps. The objective $J_{\mathrm{f}}(\pi)$ can be interpreted as penalizing the probability offset of the policy in $(\vs, \va)$ over $\pi$ and $\pi_{\mathrm{d}}$. More specifically, it aims to maximize a surrogate reward of the form $\mathcal{R}(\vs, \va) -  \beta \log\frac{\pi(\va|\vs)}{\pi_{\mathrm{d}}(\va|\vs)}$. 
However, this surrogate reward depends on the policy $\pi$, and we cannot directly apply the classical RL analyzes. 
Inspired by soft RL~\cite{haarnoja2017reinforcement,haarnoja2018soft} and offline RL~\cite{peng2019advantage}, we define our Q functions as:
\begin{equation}
\begin{split}
    Q_{\pi}(\vs, \va) = & \mathbb{E}\left[ \mathcal{R}(\vs, \va) + \sum_{i=1}^{\infty} \gamma^i \left(\mathcal{R}(\vs_i, \va_i)  - \beta\log\frac{\pi(\va_i|\vs_i)}{\pi_{\mathrm{d}}(\va_i|\vs_i)}\right) \right].
\end{split}
\end{equation}
Based on this Q function, we can simplify $J_{\mathbf{f}}$ as
\begin{equation}
\begin{split}
\label{eq_j_q}
    J_{\mathrm{f}}(\pi) = & \mathbb{E}_{\vs \sim \rho_0, \va\sim\pi}  \left[ Q_{\pi}(\vs, \va) - \beta D_{\mathrm{KL}}(\pi(\cdot|\vs) \| \pi_{\mathrm{d}}(\cdot|\vs) )\right].
\end{split}
\end{equation}
To optimize Eq.~(\ref{eq_j_q}), ExDM decouples the optimization of Q function and diffusion policy with an alternating method. In detail, we initial $\pi_0 = \pi_{\mathrm{d}}, Q_0 = Q_{\pi_0}$, then for $n=1,2,...$, we set
\begin{equation}
\begin{split}
\label{eq_alter}
    \pi_{n}(\cdot|\vs) \triangleq &\argmax_{\pi} \mathbb{E}_{\va\sim\pi}  \big[ Q_{\pi_{n-1}}(\vs, \va)  - \beta D_{\mathrm{KL}}(\pi(\cdot|\vs) \| \pi_{\mathrm{d}}(\cdot|\vs) ) \big] 
    = \frac{\pi_{\mathrm{d}} (\va|\vs) e^{ Q_{n-1}(\vs, \va) / \beta}}{Z(\vs)}, \\
    Q_{n} \triangleq & Q_{\pi_n},
\end{split}
\end{equation}
here $Z(\vs)=\int \pi_{\mathrm{d}} (\va|\vs) e^{ Q_{n}(\vs, \va) / \beta}\mathrm{d}\va$ is the partition function. 
Building on analysis of soft RL~\cite{haarnoja2017reinforcement,haarnoja2018soft}, we can demonstrate that each iteration improves the performance of the policy and that the alternating optimization will finally converge to the optimal policy of $J_{\mathrm{f}}(\pi)$.

\begin{theorem}[Proof in Appendix~\ref{app_proof_thm2}]
\label{thm_2}
ExDM can achieve policy improvement, i.e., $J_{\mathrm{f}}(\pi_{n}) \ge J_{\mathrm{f}}(\pi_{n-1})$ holds for $\forall n\ge 1$. And $\pi_n$ will converge to the optimal policy of $J_{\mathrm{f}}$.
\end{theorem}

In the following, we will introduce the practical diffusion policy fine-tuning method of ExDM for both updating the Q function and the diffusion policy, respectively, with the pseudo-code in Algorithm~\ref{algo_ExDM_finetune}.

\begin{algorithm}[t]
    \caption{Diffusion Policy Fine-tuning of ExDM} 
    \label{algo_ExDM_finetune}
    \begin{algorithmic}[1] 
        \REQUIRE Environment $\mathcal{M}$ with rewards $\mathcal{R}$, replay buffer $\mathcal{D}$, pre-trained diffusion policy $\pi_{\mathrm{d}}$ parameterized with the score model $\bm{\epsilon}_{\theta}$, fine-tuned diffusion policy $\bm{\epsilon}_{\psi}$.
        \FOR{$\text{update iteration } n = 1,2,..., N$}
        \STATE Sample $\vs$-$\va$-$\vr$ pairs $\{(\vs^m,\va^m, \vr^m)\}_{m=1}^{M}$ from $\mathcal{D}$.
        \STATE Update Q function with IQL and update Guidance $f_{\phi_{n-1}}$ with CEP.
        \STATE Optimize $\psi$ by score distillation with Eq.~(\ref{eq_score_dis}).
        \FOR{$\text{interaction step} = 1,2,..., S$}
        \STATE Interact with $\mathcal{M}$ by $\bm{\epsilon}_{\psi}$ and store state-action-reward pairs into $\mathcal{D}$.
        \ENDFOR
        \ENDFOR
    \end{algorithmic}  
\end{algorithm}

\paragraph{Q function optimization.} 
Our principle for updating Q functions is to penalize actions of which the log probability ratio between $\pi$ and $\pi_{\mathrm{d}}$ is large. Thus, we apply implicit Q-learning (IQL)~\cite{kostrikov2022offline}, which leverages expectile regression to penalize out-of-distribution actions (details are in Appendix~\ref{app_ExDM_q}).

\paragraph{Diffusion policy distillation.} 
At each update iteration $n$, given the Q function $Q_{n-1}$, it is difficult to directly calculate $\pi_{n}$ by Eq.~(\ref{eq_alter}) as $Z(s)$ is a complicated integral. 
However, sampling from $\pi_{n}$ can be regarded as sampling from an original diffusion model $\pi_{\mathrm{d}}$ with the energy guidance $Q_{n-1}$, which has been widely discussed as guided sampling~\cite{chung2022diffusion,janner2022planning,lu2023contrastive}. Especially, we employ contrastive energy prediction (CEP)~\cite{lu2023contrastive}
to sample from $\propto \pi_{\mathrm{d}} e^{Q_{n-1}/\beta}$ 
and parameterize $f_{\phi_{n-1}}(\vs, \va_t, t)$ to represent the energy guidance of timestep $t$, of which the optimization follows:
\begin{equation}
\begin{split}
    & \min_{\phi_{n-1}} \mathbb{E}_{t,\vs} \mathbb{E}_{\va^1, ..., \va^K \sim \pi_{\mathrm{d}}(\cdot|\vs)} \Bigg[
    -\sum_{i=1}^K \frac{e^{Q_{n-1}(\vs, \va^i) / \beta}}{\sum_{j=1}^K e^{Q_{n-1}(\vs, \va^j) / \beta}} \log \frac{f_{\phi_{n-1}}(\vs,\va_t^i, t)}{\sum_{j=1}^K f_{\phi_{n-1}}(\vs,\va_t^j, t)}\Bigg].
\end{split}
\end{equation}
Then ExDM fine-tunes the diffusion policy by distilling the score of $\pi_{n}$ parameterized by $\epsilon_{\psi}(\va_t|\vs, t)$:
\begin{equation}
\begin{split}
\label{eq_score_dis}
    \min_{\psi}\mathbb{E}_{\vs, \va, t} \| \bm{\epsilon}_{\psi}(\va_t | \vs, t) - \bm{\epsilon}_{\theta}(\va_t | \vs, t) -  f_{\phi_{n-1}}(\vs, \va_t, t)\|^2.
\end{split}
\end{equation}
Finally, we can directly sample from $\bm{\epsilon}_{\psi}$ to generate action of $\pi_{n}$ (details are in Appendix~\ref{app_ExDM_score}).

\section{Related Work}

\paragraph{Unsupervised Pre-training in RL.} 
To improve agent generalization, URL pre-trains agents in reward-free environments to acquire knowledge, which can later be fine-tuned to downstream tasks. 
Existing methods mainly rely on intrinsic rewards to guide agents to explore the environment, falling into two categories: exploration and skill discovery. 
Exploration methods typically explore diverse states by maximizing intrinsic rewards designed to estimate either uncertainty~\cite{pathak2017curiosity,burda2018exploration,pathak2019self,mazzaglia2022curiosity,yuan2023automatic,ying2024peac} or state entropy~\cite{lee2019efficient,liu2021behavior,liu2021aps}. 
Skill-discovery methods hope to collect diverse skills by maximizing the mutual information between skills and states~\cite{eysenbach2018diversity,lee2019efficient,campos2020explore,kim2021unsupervised,park2022lipschitz,laskin2022unsupervised,yuan2022euclid,zhao2022mixture,yang2023behavior,park2023metra,bai2024constrained,ying2024peac,wilcoxson2024leveraging}. 
Although exploring diverse states, existing methods always neglect the expression ability of pre-trained policies and choose simple Gaussian policies~\cite{pathak2017curiosity,mazzaglia2022curiosity} or skill-based policies~\cite{eysenbach2018diversity,yang2023behavior}, which fail to capture the diversity present in the explored data.
Consequently, applying generative models with strong expressive ability for improving the diversity of pre-trained policies is still less studied.

\paragraph{RL with Diffusion Models.} 
Recent advancements have shown that high-fidelity diffusion models can benefit RL from different perspectives~\cite{zhu2023diffusion}.
In offline RL, diffusion policies~\cite{wang2023diffusion,chen2023offline,chi2023diffusion,hansen2023idql,kang2024efficient} excel at modeling multimodal behaviors, outperforming previous policies such as Gaussians.
Besides policies, diffusion planners~\cite{janner2022planning,ajay2023is,he2023diffusion,liang2023adaptdiffuser,chen2024diffusion} have demonstrated the potential in long-term sequence prediction and test-time planning.
We can further refine diffusion models by energy-guided sampling~\cite{janner2022planning,lu2023contrastive,liu2024energy}.
Some works have also investigated online training diffusion policies to improve performance ~\cite{psenka2023learning,li2024learning,ren2024diffusion,mark2024policy,celik2025dime,ma2025soft,ishfaq2025langevin}. However, the computational cost of multi-step sampling remains a bottleneck for efficiency. 
In addition to behavior modeling, diffusion models have also been employed as world models~\cite{alonso2024diffusion,ding2024diffusion}, augmented replay buffer~\cite{lu2024synthetic,wang2024prioritized}, hierarchical RL~\cite{li2023hierarchical,chen2024simple}, and so on.
To the best of our knowledge, this work represents the first attempt to leverage diffusion models for unsupervised exploration, leveraging their heterogeneous data distribution modeling capabilities.

\section{Experiments}
\label{sec_expe}

In this section, we present extensive empirical results to mainly address the following questions:
\begin{itemize}
    \item Can ExDM boost the unsupervised exploration efficiency, especially in complicated mazes with numerous branching paths and decision points? (Sec.~\ref {exp_maze_exploration})
    \item What about the adaptation efficiency of the pre-trained Gaussian policies of ExDM compared to other URL baselines? (Sec.~\ref{sec_exp_finetune_gaussian})
    \item As for fast fine-tuning pre-trained diffusion policies to downstream tasks, how does the performance of ExDM compare to existing baselines? (Sec.~\ref{sec_exp_finetune_diffusion})
\end{itemize}

\begin{figure*}[t]
\centering
\includegraphics[width=0.9\linewidth]
{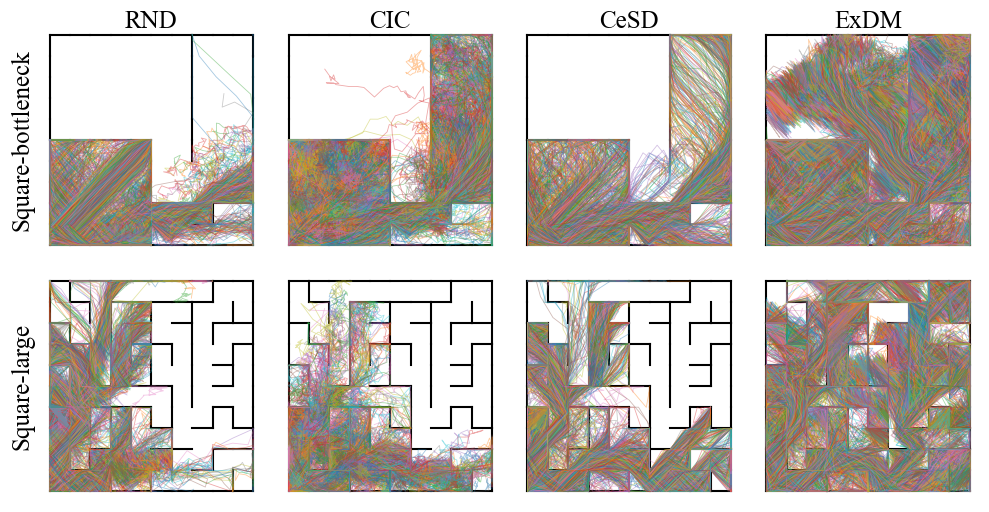}
\vspace{-1.5em}
\caption{
\textbf{Visualization of trajectories explored by different URL methods} in the most complicated mazes.
Full results of all 11 algorithms in 7 mazes are in Appendix~\ref{app_exp_maze}.
}
\label{fig_exp_replay_buffer_main}
\vspace{-1.em}
\end{figure*}

\begin{table*}[t]
\centering
\scriptsize
\resizebox{0.95\textwidth}{!}{%
\begin{tabular}{c|cccccccccccccccccc}
\toprule
Domains & Square-a & Square-b & Square-c & Square-d & Square-tree & Square-bottleneck & Square-large \\  
\midrule
ICM
& 0.58 $\pm$ 0.04 
& 0.53 $\pm$ 0.06 
& 0.47 $\pm$ 0.07
& 0.49 $\pm$ 0.06  
& 0.49 $\pm$ 0.05  
& 0.32 $\pm$ 0.07 
& 0.25 $\pm$ 0.04 \\
RND
& 0.50 $\pm$ 0.14
& 0.39 $\pm$ 0.08  
& 0.52 $\pm$ 0.16  
& 0.32 $\pm$ 0.05  
& 0.28 $\pm$ 0.06  
& 0.33 $\pm$ 0.06
& 0.33 $\pm$ 0.08 \\
Disagreement
& 0.38 $\pm$ 0.10
& 0.30 $\pm$ 0.10  
& 0.41 $\pm$ 0.19  
& 0.29 $\pm$ 0.11  
& 0.32 $\pm$ 0.11  
& 0.28 $\pm$ 0.04 
& 0.21 $\pm$ 0.06 \\
LBS
& 0.32 $\pm$ 0.04
& 0.29 $\pm$ 0.09  
& 0.27 $\pm$ 0.05  
& 0.25 $\pm$ 0.03  
& 0.22 $\pm$ 0.03  
& 0.21 $\pm$ 0.02 
& 0.19 $\pm$ 0.06 \\
\hline
DIAYN
& 0.41 $\pm$ 0.06
& 0.44 $\pm$ 0.04  
& 0.42 $\pm$ 0.04  
& 0.37 $\pm$ 0.03  
& 0.38 $\pm$ 0.06  
& 0.29 $\pm$ 0.04 
& 0.30 $\pm$ 0.04 \\
SMM
& 0.47 $\pm$ 0.13
& 0.45 $\pm$ 0.20  
& 0.36 $\pm$ 0.08  
& 0.28 $\pm$ 0.04  
& 0.25 $\pm$ 0.02
& 0.41 $\pm$ 0.13  
& 0.34 $\pm$ 0.10 \\
LSD
& 0.45 $\pm$ 0.03
& 0.38 $\pm$ 0.05  
& 0.36 $\pm$ 0.03  
& 0.35 $\pm$ 0.03  
& 0.28 $\pm$ 0.03  
& 0.34 $\pm$ 0.03  
& 0.32 $\pm$ 0.03 \\
CIC
& 0.94 $\pm$ 0.02
& \textbf{0.98 $\pm$ 0.01}  
& 0.86 $\pm$ 0.03
& 0.74 $\pm$ 0.01 
& \textbf{0.89 $\pm$ 0.01} 
& 0.58 $\pm$ 0.05 
& 0.47 $\pm$ 0.01 \\
BeCL
& 0.50 $\pm$ 0.08
& 0.48 $\pm$ 0.11  
& 0.42 $\pm$ 0.10  
& 0.37 $\pm$ 0.03  
& 0.36 $\pm$ 0.06  
& 0.29 $\pm$ 0.06 
& 0.25 $\pm$ 0.05 \\
CeSD
& 0.70 $\pm$ 0.04
& 0.79 $\pm$ 0.04  
& 0.67 $\pm$ 0.06  
& 0.46 $\pm$ 0.06  
& 0.37 $\pm$ 0.06  
& 0.46 $\pm$ 0.03 
& 0.40 $\pm$ 0.01 \\
\hline
ExDM (Ours)
& \textbf{0.99 $\pm$ 0.02}
& \textbf{0.99 $\pm$ 0.01}
& \textbf{0.98 $\pm$ 0.02}
& \textbf{0.78 $\pm$ 0.01}
& \textbf{0.91 $\pm$ 0.01}
& \textbf{0.75 $\pm$ 0.15} 
& \textbf{0.71 $\pm$ 0.07} \\
\bottomrule
\end{tabular}
}
\vspace{-0.75em}
\caption{\textbf{Detailed state coverage in Maze}. We report the mean and std of 10 seeds for each algorithm.}
\vspace{-1.em}
\label{table_maze}
\end{table*}

\subsection{Experimental Setup}

\paragraph{Maze2d.} 
We first evaluate and visualize the exploration diversity during the unsupervised stage in widely used maze2d environments~\cite{campos2020explore,yang2023behavior}. 
We choose various mazes: Square-a, Square-b, Square-c, Square-d, Square-tree, Square-bottleneck, and Square-large. 
Observations and actions in these mazes belong to $\mathbb{R}^2$. 
When interacting with the mazes, agents will be blocked when they contact the walls. 

\paragraph{Continuous Control.} 
To evaluate the fine-tuning performance in downstream tasks, we choose continuous state-based control settings in URLB~\cite{laskin2021urlb}, including 4 domains: Walker, Quadruped, Jaco, and Hopper. Each domain contains four downstream tasks. More details are in Appendix~\ref{app_exp_domain_task}.

\paragraph{Baselines.} For the \textbf{exploration in Maze2d}, we choose 4 exploration baselines: ICM~\cite{pathak2017curiosity}, RND~\cite{burda2018exploration}, Disagreement~\cite{pathak2019self}, and LBS~\cite{mazzaglia2022curiosity}; as well as 6 skill discovery baselines: DIAYN~\cite{eysenbach2018diversity}, SMM~\cite{lee2019efficient}, LSD~\cite{park2022lipschitz}, CIC~\cite{laskin2022unsupervised}, BeCL~\cite{yang2023behavior}, and CeSD~\cite{bai2024constrained}, which are standard and SOTA in this benchmark.
As for \textbf{fast fine-tuning in URLB}, we consider two settings: 
\textbf{(a)} for a fair comparison, we directly utilize DDPG~\cite{sohl2015deep} to fine-tune the pre-trained behavior Gaussian policy in ExDM, compared to existing URL baselines, including ICM, RND, Disagreement, LBS, DIAYN, SMM, LSD, CIC, BeCL, and CeSD (all baselines fine-tuned by DDPG, the standard RL backbone in URLB, except CeSD fine-tuned by ensembled DDPG);
\textbf{(b)} ExDM also fine-tunes the pre-trained diffusion policies, compared to diffusion policy fine-tuned baselines, like DQL~\cite{wang2023diffusion}, IDQL~\cite{hansen2023idql}, QSM~\cite{psenka2023learning}, and DIPO~\cite{yang2023policy}.

\paragraph{Metrics.}
In Maze2d, we pre-train agents in reward-free environments with 100k steps and visualize all collected trajectories. 
Moreover, to quantitatively compare the exploration efficiency, we evaluate the state coverage ratios, which are measured as the proportion of 0.01 $\times$ 0.01 square bins visited.
As for URLB, following standard settings, we pre-train agents in reward-free environments for 2M steps and fine-tune pre-trained policies to adapt each downstream task within extrinsic rewards for 100K steps. 
All settings are run for 10 seeds to mitigate the effectiveness of randomness.

\begin{figure*}[t]
\centering
\includegraphics[width=0.9\linewidth]
{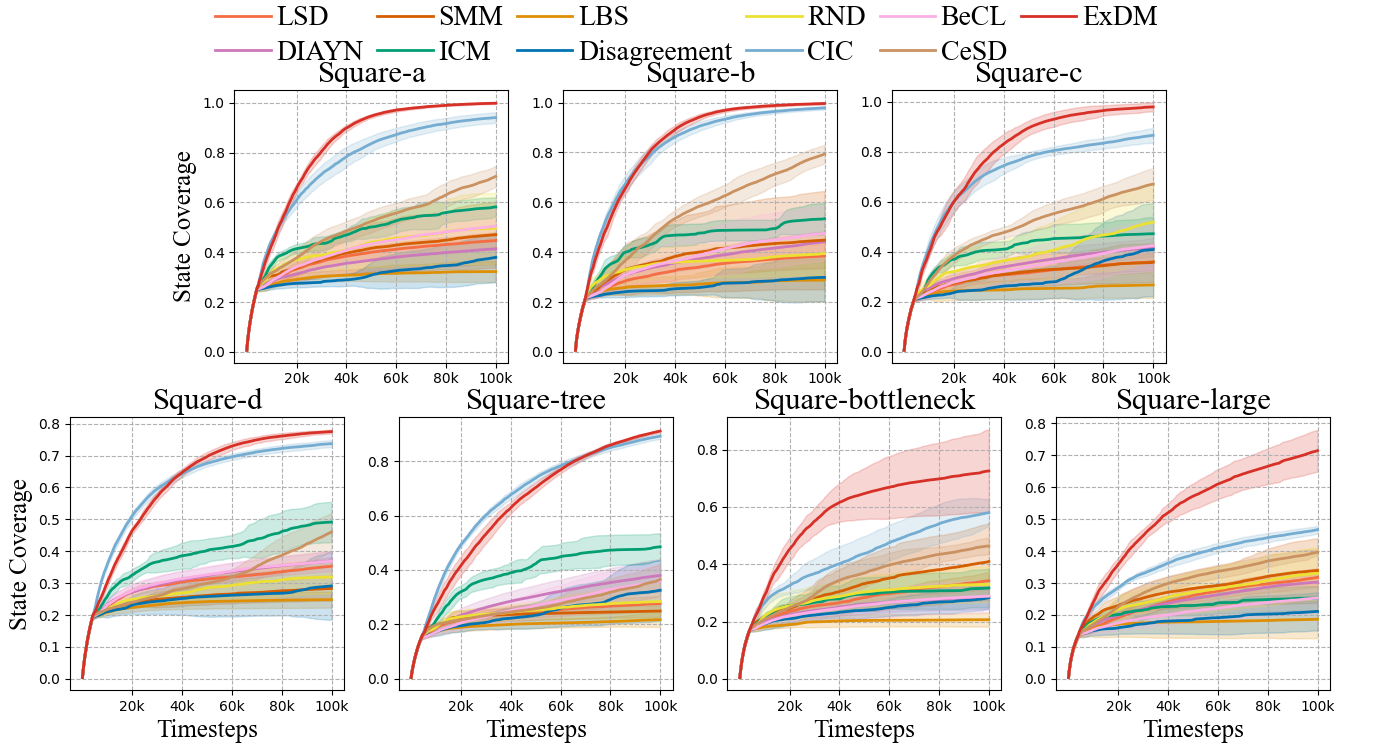}
\vspace{-1.em}
\caption{
\textbf{State coverage ratios} of different algorithms in 7 mazes during pre-training.
}
\label{fig_exp_maze_state_coverage}
\vspace{-1.em}
\end{figure*}

\begin{figure*}[t]
\centering
\resizebox{0.9\textwidth}{!}{%
\includegraphics[width=1.0\linewidth]
{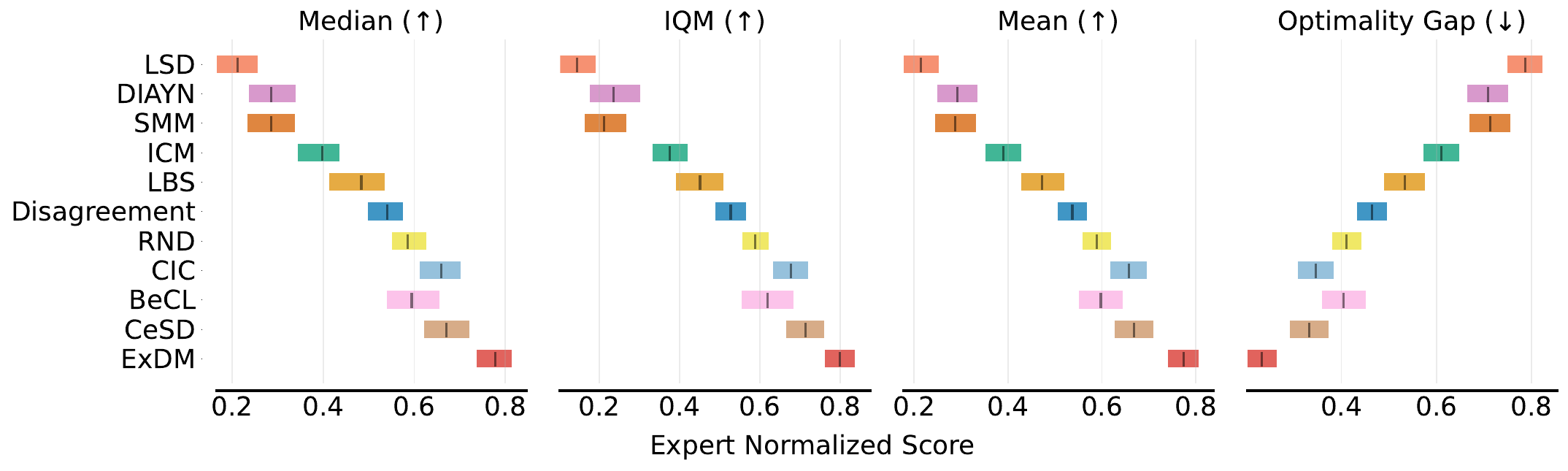}
}
\vspace{-1.em}
\caption{
\textbf{Aggregate metrics~\cite{agarwal2021deep} in URLB fine-tuned by DDPG}. Each statistic for every algorithm has 160 runs
(4 domains × 4 downstream tasks × 10 seeds).
}
\label{fig_exp_urlb}
\vspace{-1.em}
\end{figure*}

\subsection{Unsupervised Pre-training for Exploration}
\label{exp_maze_exploration}

In Fig.~\ref{fig_exp_replay_buffer_main}, we visualize trajectories collected during pre-training in complicated Square-bottleneck and Square-large (visualizations of all 7 mazes and 11 baselines are in Appendix~\ref{app_exp_maze}).
To quantitatively evaluate the exploration efficiency of each algorithm, we further plot their state coverage ratios as a function of pre-training timesteps in Fig.~\ref{fig_exp_maze_state_coverage} (final state coverages are in Table~\ref{table_maze}).
In both qualitative visualization and quantitative metrics, ExDM outperforms existing SOTA methods by large margins.
In simple mazes like Square-a, though several baselines (like CIC) can explore most states, ExDM converges much faster with better exploration efficiency.
In complicated mazes like Square-bottleneck and Square-large that have many different branching points (Fig.~\ref{fig_exp_replay_buffer_main}), all baselines will struggle at some wall corner and are unable to explore the entire maze.
In contrast, ExDM successfully explores almost the whole maze, demonstrating its strong unsupervised exploration ability.
These results demonstrate that our $\mathcal{R}_{\mathrm{score}}$, leveraging the accurate data estimation ability of diffusion models, effectively guides agents to explore more diverse states during pre-training than all baselines.

\subsection{Fine-tuning the Gaussian Policy to Downstream Tasks}
\label{sec_exp_finetune_gaussian}

We verify the ability of ExDM to fine-tune downstream tasks in URLB.
For a fair comparison, as existing URL methods directly fine-tune the pre-trained policies with DDPG, we also use DDPG to fine-tune the pre-trained Gaussian policy $\pi_{\mathrm{g}}$ in ExDM.
Following previous settings, we train DDPG agents for each downstream task with 2M steps to obtain the expert return and calculate the expert-normalized score for each algorithm.
In Fig.~\ref{fig_exp_urlb}, we compare all methods with four metrics: mean, median, interquartile mean (IQM), and optimality gap (OG), along with stratified bootstrap confidence intervals. 
As shown here, ExDM significantly outperforms all existing exploration methods (such as RND) and skill-based methods (like CIC and CeSD), demonstrating that introducing diffusion models can lead to more efficient generalization in downstream tasks.
Additional details are in Appendix~\ref{app_exp_urlb}.

\subsection{Fine-tuning the Diffusion Policy to Downstream Tasks}
\label{sec_exp_finetune_diffusion}

\begin{figure*}[t]
\centering
\resizebox{0.85\textwidth}{!}{%
\includegraphics[width=1.0\linewidth]
{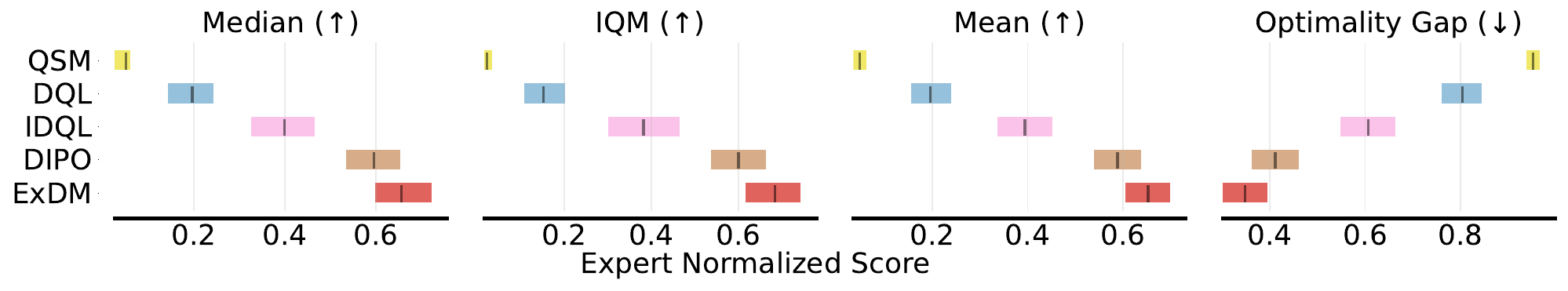}
}
\vspace{-1.em}
\caption{
Aggregate metrics~\cite{agarwal2021deep} in URLB of different fine-tuning methods for diffusion policies.
}
\label{fig_exp_urlb_diffusion}
\vspace{-1.em}
\end{figure*}

Moreover, ExDM has also pre-trained diffusion policies during unsupervised pre-training, which can capture the diversity of explored trajectories and adapt to downstream tasks.
Consequently, we compare ExDM to existing diffusion policy fine-tuning baselines.
As shown in Fig.~\ref{fig_exp_urlb_diffusion}, ExDM substantially outperforms existing baselines, demonstrating the efficiency of its alternating optimization.
However, the diffusion policy fine-tuning performance is still lower than the Gaussian policy performance, which may be due to the limited interaction timestep during the online fine-tuning stage.
It is an interesting future direction to design more efficient diffusion policy online fine-tuning methods.

\begin{figure}[t]
\subfigure[]{
\begin{minipage}[t]{0.352\linewidth}
\centering
\includegraphics[width=0.9\linewidth]{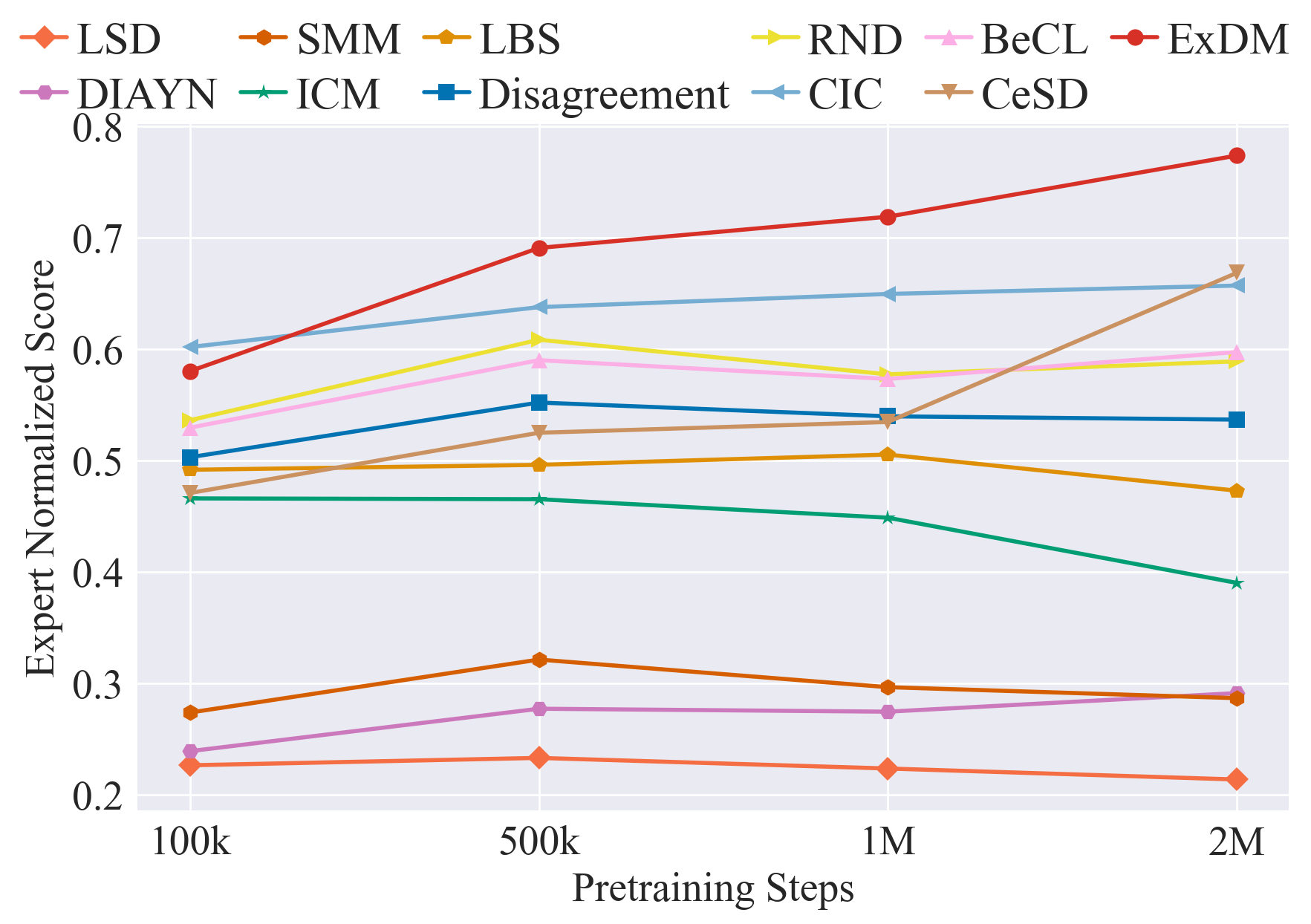}
\label{fig_exp_timestep_ablation}
\end{minipage}}
\subfigure[]{
\begin{minipage}[t]{0.264\linewidth}
\centering
\includegraphics[width=0.9\linewidth]{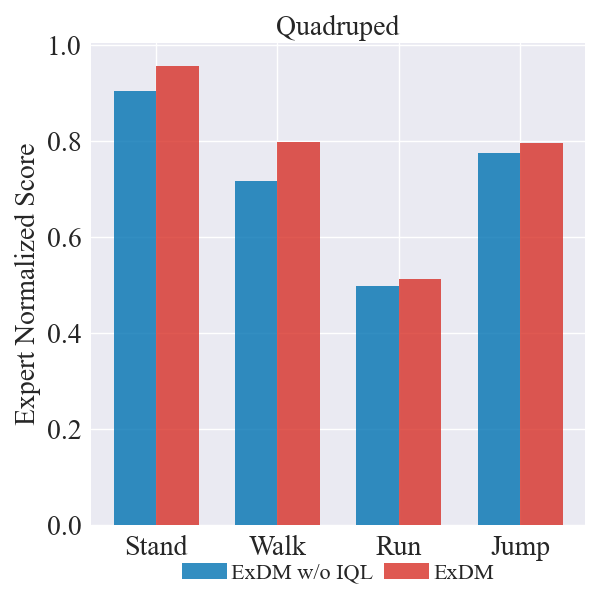}
\label{fig_exp_ablation}
\end{minipage}}
\subfigure[]{
\begin{minipage}[t]{0.352\linewidth}
\centering
\includegraphics[width=0.9\linewidth]{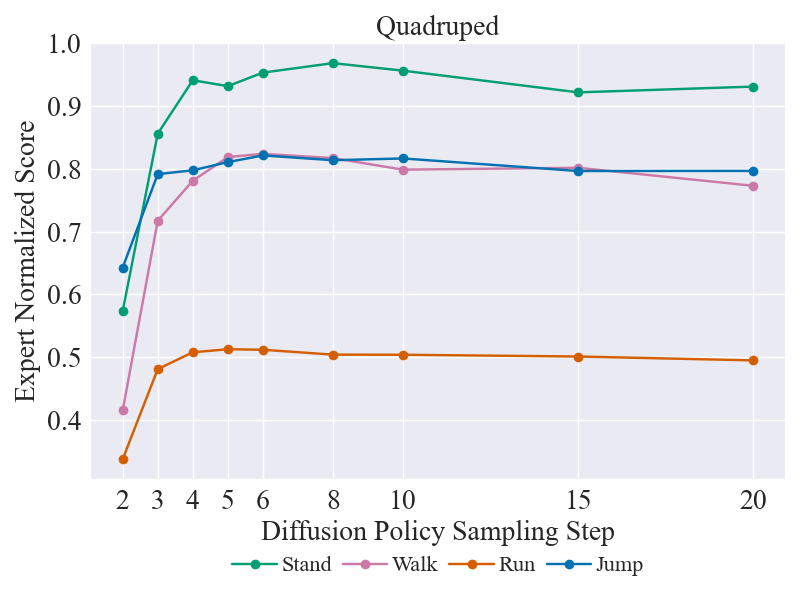}
\label{fig_exp_ablation_sampling_step}
\end{minipage}}
\vspace{-1.25em}
\caption{
Ablation studies on
\textbf{(a)} pre-training timesteps;
\textbf{(b)} Q learning choices
(ExDM w/o IQL utilizes in-sample Q learning);
\textbf{(c)} fine-tuned diffusion policy sampling steps for downstream tasks.
}
\label{minigrid_fig}
\vspace{-1.25em}
\end{figure}



\subsection{Ablation Studies}
\label{sec_abla}

\paragraph{Pre-training Steps.}
We first do ablation studies on pre-trained steps (100k, 500k, 1M, and 2M) to evaluate fine-tuned performance (100k fine-tuned steps).
As shown in Fig.~\ref{fig_exp_timestep_ablation}, ExDM markedly exceeds all baselines from 500k steps, indicating that the diffusion model enhances fine-tuning. 
Moreover, ExDM substantially improves with pre-training timesteps increasing, showing that the unsupervised exploration benefits downstream tasks. 
Additional results are in Appendix~\ref{app_exp_timestep_results}.


\paragraph{Q function optimization.} 
We conduct ablation studies to evaluate the impact of different Q learning methods during fine-tuning.
In detail, we introduce ExDM w/o IQL, which utilizes In-support Softmax Q-Learning~\cite{lu2023contrastive} for optimizing Q functions.
Results in Fig.~\ref{fig_exp_ablation} demonstrate that ExDM consistently outperforms ExDM w/o IQL, verifying the efficiency of IQL in diffusion policy fine-tuning.

\paragraph{Sampling steps of diffusion policies.} 
During the diffusion policy fine-tuning, ExDM requires sampling actions from diffusion policies for both trajectory generation and final evaluation. 
We adopt DPM-Solver~\cite{lu2022dpm} to accelerate the sampling.
For trajectory collection, we set the diffusion step to 15, following previous works~\cite{lu2023contrastive}.
Then we conduct an ablation study between the diffusion sampling steps used during inference and the fine-tuned policy performance.
Fig.~\ref{fig_exp_ablation_sampling_step} shows that performance improves as the diffusion steps increase and gradually stabilizes when the sampling step exceeds 5.

\subsection{Limitation and Discussion}
\label{sec_limitation}

ExDM employs diffusion policies to fit heterogeneous explored data, which is a natural fit for continuous control.
However, applying diffusion policies in tasks with discrete action spaces remains a challenge.
Moreover, ExDM applies diffusion models to boost unsupervised exploration, which may require more computational resources, leading to a trade-off between efficiency and performance. 

\section{Conclusion}
\label{sec_con}

Unsupervised exploration is one of the major problems in RL for improving task generalization, as it relies on accurate intrinsic rewards to guide exploration of unseen regions.
In this work, we address the challenge of limited policy expressivity in previous exploration methods by leveraging the powerful expressive ability of diffusion policies.
In detail, our Exploratory Diffusion Model (ExDM) improves exploration efficiency during pre-training while generating policies with high behavioral diversity.
We also provide a theoretical analysis of diffusion policy fine-tuning, along with practical alternating optimization methods.
Experiments in various settings demonstrate that ExDM can effectively benefit both pre-training exploration and fine-tuning performance.
We hope this work can inspire further research in developing high-fidelity generative models for improving unsupervised exploration, particularly in large-scale pre-trained agents or real-world control applications.

\newpage
{
    \small
    \bibliographystyle{plain}
    \bibliography{ref}
}

\newpage
\appendix

\section{Theoretical analyses}
\label{app_proof}

\subsection{The details and proof of Theorem~\ref{thm_1}}
\label{app_proof_thm_1}

In this part, we narrate Theorem~\ref{thm_1} in detail as well as provide its proof. 

Assume that $\mathcal{S}, \mathcal{A}$ are discrete spaces and $|\mathcal{S}| = S, |\mathcal{A}| = A$. For any policy $\pi:\mathcal{S}\rightarrow\Delta(\mathcal{A})$, we define the discount state distribution of $\pi$ as $d_{\pi}(\vs) = (1-\gamma)\sum_{t=0}^{\infty} \left[ \gamma^t \mathcal{P}(\vs^t=\vs)\right]$. Naturally, $d_{\pi}(\cdot)$ is a distribution of the state space $\mathcal{S}$. Specially, when $\mathcal{S}$ is discrete, following previous work~\cite{eysenbach2021information,ying2024peac}, we can regard $d_{\pi}$ as a point of the probability simplex in $\mathbb{R}^S$, i.e., $d_{\pi} \in H \triangleq \{(x_i)_{i=1}^S | \sum_{i=1}^S x_i = 1, 0\leq x_i \leq 1\}$ (for example, the light blue plane in the left part of Fig.~\ref{fig_overview}).
Moreover, we consider $D \triangleq \{d_{\pi} \in H | \forall \pi\} \subseteq H$ representing all feasible state distribution. 

It is natural that there are $M\triangleq A^S$ different deterministic policies. 
Previous work~\cite{eysenbach2021information} has proven that $D$ is a convex polytope, of which the vertices are the state distributions of the deterministic policies (for example, the green points of Fig.~\ref{fig_overview}).
Consequently, for any downstream task represented by some extrinsic reward function $\mathcal{R}$, the optimal policy is one the vertices of $D$, i.e., some deterministic policy.

Differently, the unsupervised exploration in URL aims to maximize the entropy of the policy, i.e., we hope to optimize $\hat{\pi} = \argmax_{\pi} \mathcal{H}(d_{\pi}(\cdot))$. As $\mathcal{H}(d_{\pi}(\cdot)) = \int_{\vs \sim d_{\pi}(\cdot)} [-\log d_{\pi}(\vs)] d\vs$, this problem can be regard as maximizing a surrogate reward $-\log d_{\pi}(\vs)$. However, the surrogate reward is related to the current policy $\pi$, thus the analyses of standard RL may not hold.
Actually, if we consider all distribution over $\mathcal{S}$, i.e., all points in $H$, it is well known that the distribution with the maximal distribution is the center of $H$, i.e., $O=(1/S, 1/S, ..., 1/S)\in H$ (for example, the red point of Fig.~\ref{fig_overview}).
Although $O\in D$ may not hold, we hope to claim that the optimal policy $\hat{\pi}$ with the maximal state distribution entropy may not be deterministic, and its state distribution is $O$, i.e., $O\in D$, with high probability.

We consider the distribution in which the state distributions of the $M$ deterministic policies are i.i.d., and all follow the uniform distribution on $H$.
Therefore, our problem can be transformed into: \emph{there are $M$ i.i.d. points uniformly sampled from $H$, and the convex polytope formed by these $M$ points is $D \subseteq H$, then calculating the probability of the event $O\in D$}.

Based on the results in geometric probability (Eq.~(13) of paper~\cite{baddeley2007random}, extending the Wendel Theorem in~\cite{wendel1962problem}), we have
\begin{equation}
\begin{split}
    P(O\in D) \ge & 1 - \sum_{k=0}^{S-1} C_{M}^{k}\left(\frac{u(O)}{2}\right)^k \left(1- \frac{u(O)}{2}\right)^{M-k},
\end{split}
\end{equation}
here $u(O) = \mathrm{vol} ( (2O - H) \cap H) / \mathrm{vol} (H)$. Below we first estimate $u(O)$, which is obviously belong to $[0, 1]$. We have
\begin{equation}
\begin{split}
    \mathrm{vol} (H) = & \sqrt{S} \int_{x_1+x_2+...+x_S = 1, x_i \ge 0} dx_1 dx_2 ... dx_S \\
    = & \sqrt{S} \int_{0}^1 dx_1 \int_{0}^{1-x_1} dx_2 ... \int_{0}^{1-x_1-x_2-...-x_{S-2}} dx_{S-1} \\
    = & \sqrt{S} \int_{0}^1 dx_1 \int_{0}^{1-x_1} dx_2 ... \int_{0}^{1-x_1-x_2-...-x_{S-3}} (1-x_1-...-x_{S-2}) dx_{S-2} \\
    = & \sqrt{S} \int_{0}^1 dx_1 \int_{0}^{1-x_1} dx_2 ... \int_{0}^{1-x_1-x_2-...-x_{S-4}} \frac{(1-x_1-...-x_{S-3})^2}{2} dx_{S-3} \\
    = & ... \\
    = & \sqrt{S} \int_{0}^1  \frac{(1-x_1)^{S-2}}{(S-2)!}  dx_{1} \\
    = & \frac{\sqrt{S}}{(S-1)!}.
\end{split}
\end{equation}
Then we estimate $\mathrm{vol} ( (2O - H) \cap H)$.
$H$ is a convex polytope surrounded by $S$ points $A_1 = (1, 0, 0, ..., 0), ..., A_S = (0, 0, ..., 0, 1)$.
As $O = (1/S, 1/S, ..., 1/S)$, $2O - H$ is a convex polytope surrounded by $S$ points $B_1 = (2/S - 1, 1/S, 1/S, ..., 1/s), ..., B_S = (1/S, 1/S, ..., 1/S, 2/S - 1)$.
It is difficult to directly calculate the volume of $(2O-H) \cap H$ (although we can calculate it by the inclusion-exclusion principle).
We can provide a lower bound of $(2O-H) \cap H$.
Consider a convex polytope $C$ surrounded by $S$ points $C_1 = (0, 1/(S-1), 1/(S-1), ..., 1/(S-1)), ..., C_S = (1/(S-1), 1/(S-1), ..., 1/(S-1), 0)$. It is easy to show that $C \subseteq (2O-H) \cap H$ and $C\sim H$. As the side length of $H$ and $C$ is $\sqrt{\frac{2}{S}}$ and $\sqrt{\frac{2}{S}}\frac{1}{S-1}$. Thus we have $\mathrm{vol} ( (2O - H) \cap H) \ge \mathrm{vol}(C) = \frac{1}{(S-1)^S} \frac{\sqrt{S}}{(S-1)!}$ and $u(O) = \mathrm{vol} ( (2O - H) \cap H) / \mathrm{vol} (H) \ge \frac{1}{(S-1)^S}$. Assume that $M \ge S - 2 + (S-1)\frac{2 - u(O)}{u(O)}$, then for $\forall  0 \leq k \leq S-2$, we have
\begin{equation}
\begin{split}
    & C_{M}^k \left(\frac{u(O)}{2}\right)^k \left(1- \frac{u(O)}{2}\right)^{M-k} \\
    = & \frac{k+1}{M-k} \frac{2- u(O)}{u(O)} C_{M}^{k+1} \left(\frac{u(O)}{2}\right)^{k+1} \left(1- \frac{u(O)}{2}\right)^{M-k-1} \\
    \leq & \frac{S-2+1}{M-S+2} \frac{2- u(O)}{u(O)} C_{M}^{k+1} \left(\frac{u(O)}{2}\right)^{k+1} \left(1- \frac{u(O)}{2}\right)^{M-k-1} \\
    \leq & \frac{S-1}{(S-1)\frac{2 - u(O)}{u(O)}} \frac{2- u(O)}{u(O)} C_{M}^{k+1} \left(\frac{u(O)}{2}\right)^{k+1} \left(1- \frac{u(O)}{2}\right)^{M-k-1} \\
    = & C_{M}^{k+1} \left(\frac{u(O)}{2}\right)^{k+1} \left(1- \frac{u(O)}{2}\right)^{M-k-1}.
\end{split}
\end{equation}
Consequently, we set $v(S) = 1- \frac{u(O)}{2} < 1$ and have
\begin{equation}
\begin{split}
    P(O\in D) \ge & 1 - \sum_{k=0}^{S-1} C_{M}^{k}\left(\frac{u(O)}{2}\right)^k \left(1- \frac{u(O)}{2}\right)^{M-k}\\
    \ge & 1 -  C_{M}^{S-1} S \left(\frac{u(O)}{2}\right)^{S-1} \left(1- \frac{u(O)}{2}\right)^{M-S+1} \\
    = & 1 -  C_{M}^{S-1} S \left(\frac{u(O)}{2 - u(O)}\right)^{S-1} \left(1- \frac{u(O)}{2}\right)^{M} \\
    \ge & 1 -  C_{M}^{S-1} S v(S)^{M} \\
    = & 1 -  \frac{M\times (M-1)\times ...\times (M-S+2)}{1\times 2\times ...\times (S-1)} S v(S)^{M} \\
    \ge & 1 -  M\times (M-1)\times ...\times (M-S+2)\times S v(S)^{M} \\
    \ge & 1 - M^{S} v(S)^M.
\end{split}
\end{equation}
If we fix $S$, with the increasing of $A$, $M^{S} v(S)^M$ will fast converge to 0 as $0< v(S) < 1$.
Thus, we have proven Theorem~\ref{thm_1}.

\newpage

\subsection{The proof of Theorem~\ref{thm_2}}
\label{app_proof_thm2}

Below, we first analyze our fine-tuning objective Eq.~\ref{eq_finetune_obj} and then prove Theorem~\ref{thm_2}.

Assuming $\rho_0$ is the original state distribution of the MDP $\mathcal{M}$, we have 
\begin{equation}
\begin{split}
    J_{\mathrm{f}}(\pi) \triangleq & J(\pi) - \frac{\beta}{1-\gamma} \mathbb{E}_{\vs\sim d_{\pi}}  \left[D_{\mathrm{KL}}(\pi(\cdot|\vs) \| \pi_{\mathrm{d}}(\cdot|\vs) ) \right] \\
    = & \frac{1}{1-\gamma} \mathbb{E}_{\vs\sim d_{\pi}, \va \sim \pi(\cdot|\vs)}  \left[\mathcal{R}(\vs, \va) -  \beta D_{\mathrm{KL}}(\pi(\cdot|\vs) \| \pi_{\mathrm{d}}(\cdot|\vs) ) \right] \\
    = & \mathbb{E}_{\vs \sim \rho_0, \va \sim \pi(\cdot|\vs)}  \left[ \sum_{i=0}^{\infty} \gamma^i \left(\mathcal{R}(\vs_i, \va_i) -  \beta D_{\mathrm{KL}}(\pi(\cdot|\vs_i) \| \pi_{\mathrm{d}}(\cdot|\vs_i) ) \right) \Bigg| \vs_0=\vs, \va_0=\va\right]\\
    = & \mathbb{E}_{\vs \sim \rho_0, \va \sim \pi(\cdot|\vs)}  \Bigg[ \mathcal{R}(\vs, \va) - \beta D_{\mathrm{KL}}(\pi(\cdot|\vs) \| \pi_{\mathrm{d}}(\cdot|\vs) ) \\
    & \qquad \qquad \qquad + \sum_{i=1}^{\infty} \gamma^i \left(\mathcal{R}(\vs_i, \va_i) -  \beta D_{\mathrm{KL}}(\pi(\cdot|\vs_i) \| \pi_{\mathrm{d}}(\cdot|\vs_i) ) \right) \Bigg] \\
    = & \mathbb{E}_{\vs \sim \rho_0, \va \sim \pi(\cdot|\vs)}  \left[ Q_{\pi}(\vs, \va) - \beta D_{\mathrm{KL}}(\pi(\cdot|\vs) \| \pi_{\mathrm{d}}(\cdot|\vs) )\right],
\end{split}
\end{equation}
here we set
\begin{equation}
\begin{split}
    Q_{\pi}(\vs, \va) & = \mathbb{E}\left[ \mathcal{R}(\vs, \va) + \sum_{i=1}^{\infty} \gamma^i \left(\mathcal{R}(\vs_i, \va_i)  - \beta\log\frac{\pi(\va_i|\vs_i)}{\pi_{\mathrm{d}}(\va_i|\vs_i)}\right) \right] \\
    & = \mathbb{E}\left[ \mathcal{R}(\vs, \va) + \sum_{i=1}^{\infty} \gamma^i \left(\mathcal{R}(\vs_i, \va_i) -  \beta D_{\mathrm{KL}}(\pi(\cdot|\vs_i) \| \pi_{\mathrm{d}}(\cdot|\vs_i) ) \right)  \right].
\end{split}
\end{equation}
As discussed in Sec.~\ref{sec_finetune}, ExDM applies the following alternative optimization method:
\begin{equation}
\begin{split}
    \pi_{n}(\cdot|\vs) = &\argmax_{\pi} \mathbb{E}_{\va\sim\pi(\cdot|\vs)} [ Q_{\pi_{n-1}}(\vs, \va)  - \beta D_{\mathrm{KL}}(\pi(\cdot|\vs) \| \pi_{\mathrm{d}}(\cdot|\vs) ) ], \\
    Q_{n} = & Q_{\pi_n},
\end{split}
\end{equation}
Now we prove that $\pi_n(\va|\vs) = \frac{1}{Z(\vs)} \pi_{\mathrm{d}} (\va|\vs) e^{Q_{n-1}(\vs, \va) / \beta}$. More generally, we define 
\begin{equation}
\begin{split}
    F(\pi, \pi', \vs) = \mathbb{E}_{\va\sim\pi(\cdot|\vs)}  \left[ Q_{\pi'}(\vs, \va) - \beta D_{\mathrm{KL}}(\pi(\cdot|\vs) \| \pi_{\mathrm{d}}(\cdot|\vs) )\right].
\end{split}
\end{equation}
Using the calculus of variations, we can calculate the optimal point $\pi^*$ of $F$ satisfying that 
\begin{equation}
\begin{split}
    Q_{\pi'}(\vs, \va) = \beta  \log\frac{\pi^*(\va|\vs)}{\pi_{\mathrm{d}}(\va|\vs)} + b \beta,
\end{split}
\end{equation}
here $b$ is a constant not related to $\pi^*$, and we have $\pi^*(\va|\vs) = \pi_{\mathrm{d}}(\va|\vs) e^{\frac{Q_{\pi'}(\vs, \va)}{\beta} - b}$. As $\int \pi^*(\va|\vs) d\va = 1$, we can calculate that
\begin{equation}
\begin{split}
    b =& \log\int \pi_{\mathrm{d}}(\va|\vs) e^{\frac{Q_{\pi'}(\vs, \va)}{\beta}} d\va, \quad 
    \pi^*(\va|\vs) =  \frac{\pi_{\mathrm{d}}(\va|\vs) e^{\frac{Q_{\pi'}(\vs, \va)}{\beta} - b}}{\int \pi_{\mathrm{d}}(\va|\vs) e^{\frac{Q_{\pi'}(\vs, \va)}{\beta}} d\va}.
\end{split}
\end{equation}
i.e., we have $\argmax_{\pi} F(\pi, \pi', \vs) \propto \pi_{\mathrm{d}} (\cdot|\vs) e^{Q_{\pi'}(\vs, \cdot) / \beta}$ and thus $\pi_n(\va|\vs) = \frac{1}{Z(\vs)} \pi_{\mathrm{d}} (\va|\vs) e^{Q_{n-1}(\vs, \va) / \beta}$. 

Below we will prove Theorem~\ref{thm_2}.
\begin{proof}
Based on the definition of $F$, we have $J_{\mathrm{f}}(\pi) = \mathbb{E}_{\vs\sim\rho_0}F(\pi, \pi, \vs)$. Thus we require to prove $\mathbb{E}_{\vs\sim\rho_0}F(\pi_n, \pi_n, \vs) \ge \mathbb{E}_{\vs\sim\rho_0}F(\pi_{n-1}, \pi_{n-1}, \vs)$. As we have discussed above,\begin{equation}
\begin{split}
    \pi_{n}(\cdot|\vs)
    = & \argmax_{\pi} F(\pi, \pi_{n-1}, \vs) = \frac{1}{Z(\vs)} \pi_{\mathrm{d}} (\va|\vs) e^{ Q_{\pi_{n-1}}(\vs, \va)/ \beta}. \\
    F(\pi_{n}, \pi_{n-1}, \vs) \ge & F(\pi_{n-1}, \pi_{n-1}, \vs).
\end{split}
\end{equation}


In other words, we have proven that $\mathbb{E}_{\vs\sim\rho_0}F(\pi_n, \pi_{n-1}, \vs) \ge \mathbb{E}_{\vs\sim\rho_0}F(\pi_{n-1}, \pi_{n-1}, \vs)$. Moreover, we have
\begin{equation}
\begin{split}
    Q_{\pi_{n-1}}(\vs, \va) 
    = &\mathcal{R}(\vs, \va) + \mathbb{E}\left[  \sum_{i=1}^{\infty} \gamma^i \left(\mathcal{R}(\vs_i, \va_i)  - \beta D_{\mathrm{KL}} (\pi_{n-1}(\cdot|\vs_i) \| \pi_{\mathrm{d}}(\cdot|\vs_i) ) \right) \Bigg{|} \vs_0=\vs, \va_0=\va \right] \\
    = &\mathcal{R}(\vs, \va) - \beta \gamma \mathbb{E}_{\vs_1} \left(D_{\mathrm{KL}} (\pi_{n-1}(\cdot|\vs_1) \| \pi_{\mathrm{d}}(\cdot|\vs_1) )\right) + \gamma \mathbb{E}_{\vs_1,\va_1} \left[ Q_{\pi_{n-1}}(\vs_1, \va_1)\right] \\
    = &\mathcal{R}(\vs, \va) + \gamma \mathbb{E}_{\vs_1} F(\pi_{n-1}, \pi_{n-1}, \vs).
\end{split}
\end{equation}
Thus
\begin{equation}
\begin{split}
    & Q_{\pi_{n}}(\vs, \va) - Q_{\pi_{n-1}}(\vs, \va)\\
    = & \gamma\mathbb{E}_{\vs_1} \left[ F(\pi_{n}, \pi_{n}, \vs_1) - F(\pi_{n-1}, \pi_{n-1}, \vs_1)\right]
    \ge \gamma\mathbb{E}_{\vs_1} \left[ F(\pi_{n}, \pi_{n}, \vs_1) - F(\pi_{n}, \pi_{n-1}, \vs_1)\right] \\
    = & \gamma\mathbb{E}_{\vs_1} \mathbb{E}_{\va_1\sim\pi_{n}} [ Q_{\pi_{n}}(\vs_1, \va_1) - \beta D_{\mathrm{KL}}(\pi_{n}(\cdot|\vs_1) \| \pi_{\mathrm{d}}(\cdot|\vs_1)\\
    & \qquad \qquad \quad - Q_{\pi_{n-1}}(\vs_1, \va_1) + \beta D_{\mathrm{KL}}(\pi_{n}(\cdot|\vs_1) \| \pi_{\mathrm{d}}(\cdot|\vs_1) )] \\
    = & \gamma\mathbb{E}_{\vs_1}\mathbb{E}_{\va_1\sim\pi_{n}}  \left[ Q_{\pi_{n}}(\vs_1, \va_1) - Q_{\pi_{n-1}}(\vs_1, \va_1) \right].
\end{split}
\end{equation}
Given the property of $d_{\pi}$ that $d_{\pi}(\vs) - (1-\gamma)\rho_0(\vs) = \gamma\sum_{\vs'}d_{\pi}(\vs')\sum_{\va}\pi(\va|\vs')\mathcal{P}(\vs|\vs', \va)$~\cite{ying2022towards}, we have
\begin{equation}
\begin{split}
    &\mathbb{E}_{\vs\sim d_{\pi_{n}}, \va\sim\pi_{n}(\cdot|\vs)} \left[Q_{\pi_{n}}(\vs, \va) - Q_{\pi_{n-1}}(\vs, \va)\right] \\
    \ge & \gamma \mathbb{E}_{\vs\sim d_{\pi_{n}}, \va\sim\pi_{n}(\cdot|\vs)}\mathbb{E}_{\vs_1} \mathbb{E}_{\va_1\sim\pi_{n}}  \left[ Q_{\pi_{n}}(\vs_1, \va_1) - Q_{\pi_{n-1}}(\vs_1, \va_1) \right] \\
    = & \int \left(d_{\pi_n}(\vs_1) - (1-\gamma)\rho_0(\vs_1)\right) \mathbb{E}_{\va_1\sim\pi_{n}} \left[ Q_{\pi_{n}}(\vs_1, \va_1) - Q_{\pi_{n-1}}(\vs_1, \va_1) \right] d\vs_1\\
    = & \mathbb{E}_{\vs_1\sim d_{\pi_{n}}, \va_1\sim\pi_{n}(\cdot|\vs_1)}  \left[ Q_{\pi_{n}}(\vs_1, \va_1) -  Q_{\pi_{n-1}}(\vs_1, \va_1) \right] \\
    - & (1-\gamma ) \mathbb{E}_{\vs_1\sim \rho_0, \va_1\sim\pi_{n}(\cdot|\vs_1)}  \left[ Q_{\pi_{n}}(\vs_1, \va_1) - Q_{\pi_{n-1}}(\vs_1, \va_1) \right].
\end{split}
\end{equation}
Consequently,
\begin{equation}
\begin{split}
    & \mathbb{E}_{\vs\sim \rho_0} F(\pi_{n}, \pi_{n-1}, \vs) -  \mathbb{E}_{\vs\sim \rho_0} F(\pi_{n-1}, \pi_{n-1}, \vs)\\
    = &\mathbb{E}_{\vs_1\sim \rho_0,\va_1\sim\pi_{n}(\cdot|\vs_1)}  \left[ Q_{\pi_{n}}(\vs_1, \va_1) - Q_{\pi_{n-1}}(\vs_1, \va_1) \right] \ge 0
\end{split}
\end{equation}
Finally, we have 
\begin{equation}
\begin{split}
    J_{\mathrm{f}}(\pi_{n}) = \mathbb{E}_{\vs\sim \rho_0} F(\pi_{n}, \pi_{n}, \vs) \ge \mathbb{E}_{\vs\sim \rho_0} F(\pi_{n}, \pi_{n-1}, \vs) \ge \mathbb{E}_{\vs\sim \rho_0} F(\pi_{n-1}, \pi_{n-1}, \vs) = J_{\mathrm{f}}(\pi_{n-1}).
\end{split}
\end{equation}
Thus, our policy iteration can improve the performance. Moreover, under some regularity conditions, $\pi_n$ converges to $\pi_{\infty}$. Since non-optimal policies can be improved by our iteration, the converged policy $\pi_{\infty}$ is optimal for $J_{\mathrm{f}}$.
\end{proof}

\newpage
\section{Details of ExDM}
\label{app_alog}

Below we discuss more details about the Q function optimization and diffusion policy optimization of ExDM during the fine-tuning stage.

\subsection{Q function optimization}
\label{app_ExDM_q}

For the Q function optimization, we choose to use implicit Q-learning (IQL)~\cite{kostrikov2022offline}, which is efficient to penalize out-of-distribution actions~\cite{hansen2023idql}. The main training pipeline of IQL is expectile regression, i.e.,
\begin{equation}
\begin{split}
    &\min_{\zeta} L_{V}(\zeta) = \mathbb{E}_{\vs,\va\sim\mathcal{D}} \left[L_2^{\tau} (Q_{\phi}(\vs, \va) - V_{\psi}(\vs))\right],\\
    & \min_{\phi} L_{Q}(\phi) = \mathbb{E}_{\vs,\va,\vs'\sim\mathcal{D}} \left[\| r(\vs, \va) + \gamma V_{\zeta}(\vs') - Q_{\phi}(\vs, \va)\|^2 \right],
\end{split}
\end{equation}
here $L_2^{\tau}(\vu) = |\tau - \mathds{1}(\vu<0)|\vu^2$ and $\tau$ is a hyper-parameter. In detail, when $\tau > 0.5$, $L_2^{\tau}$ will downweight actions with low Q-values and give more weight to actions with larger Q-values.

\subsection{Diffusion Policy Fine-tuning}
\label{app_ExDM_score}

For sampling from $\pi_{n} =  \frac{1}{Z(s)} \pi_{\mathrm{d}} e^{Q_{n-1}/\beta}$, we choose contrastive energy prediction (CEP)~\cite{lu2023contrastive}, a powerful guided sampling method. First, we calculate the score function of $\pi_n$ as
\begin{equation}
    \nabla_{\va} \log \pi_{n}(\va|\vs) = \nabla_{\va} \log \pi_{\mathrm{d}}(\va|\vs) + \frac{1}{\beta} \nabla_{\va} Q_{n-1}(\vs, \va).
\end{equation}
Moreover, to calculate the score function of $\pi_n$ at each timestep $t$, i.e., $\nabla_{\va_t}\log \pi_t^n(\va|\vs)$, CEP further defines the following Intermediate Energy Guidance:
\begin{equation}
    \mathcal{E}_t^{n-1} (\vs, \va_t) = \left\{
    \begin{aligned}
    &\frac{1}{\beta} Q_{n-1}(\vs, \va_0), & t=0 \\
    &\log \mathbb{E}_{\mu_{0t}(\va_0|\vs, \va_t)}\left[e^{ Q_{n-1}(\vs, \va_0) / \beta}\right], & t>0 
    \end{aligned}
    \right.
\end{equation}
Then Theorem 3.1 in CEP proves that
\begin{equation}
\begin{split}
    \pi_{t}^{n} (\va_t|\vs) \propto & \pi_{\mathrm{d}}(\va_t|\vs) e^{\mathcal{E}_t^{n-1}(\vs,\va_t)}, \\
    \nabla_{\va_t} \log \pi_{t}^{n}(\va_t|\vs) = & \nabla_{\va_t} \log \pi_{\mathrm{d}}(\va_t|\vs) + \nabla_{\va} \mathcal{E}_t^{n-1}(\vs, \va_t).
\end{split}
\end{equation}
For estimating $\nabla_{\va} \mathcal{E}_t^{n-1}(\vs, \va_t)$, CEP considers a parameterized neural network $f_{\phi_{n-1}}(\vs, \va_t, t)$ with the following objective:
\begin{equation}
\begin{split}
    & \min_{\phi_{n-1}} \mathbb{E}_{t,\vs} \mathbb{E}_{\va^1, ..., \va^K \sim \pi_{\mathrm{d}}(\cdot|\vs)} \Bigg[ -\sum_{i=1}^K \frac{e^{Q_{n-1}(\vs, \va^i) / \beta}}{\sum_{j=1}^K e^{Q_{n-1}(\vs, \va^j) / \beta}} \log \frac{f_{\phi_{n-1}}(\vs,\va_t^i, t)}{\sum_{j=1}^K f_{\phi_{n-1}}(\vs,\va_t^j, t)}\Bigg].
\end{split}
\end{equation}
Then Theorem 3.2 in CEP~\cite{lu2023contrastive} has proven that its optimal solution $f_{\phi_{n-1}^*}$ satisfying that $\nabla_{\va_t} f_{\phi_{n-1}^*}(\vs, \va_t, t) = \nabla_{\va_t} \mathcal{E}_t^{n-1}(\vs, \va_t)$.

Consequently, we propose to fine-tune $\nabla_{\va_t} \log \pi_{t}^{n}(\va_t|\vs)$ parameterized as $\vs_{\psi}(\va_t | \vs, t)$ with the following distillation objective:
\begin{equation}
\begin{split}
    \min_{\psi}\mathbb{E}_{\vs, \va, t} \| \bm{\epsilon}_{\psi}(\va_t | \vs, t) - \bm{\epsilon}_{\theta}(\va_t | \vs, t) -  f_{\phi_{n-1}}(\vs, \va_t, t)\|^2.
\end{split}
\end{equation}
And the optimal solution $\psi^*$ satisfying that $\bm{\epsilon}_{\psi^*}(\va_t | \vs, t)$ is the score function of $\pi_n$, i.e., we can sample from $\bm{\epsilon}_{\psi^*}(\va_t | \vs, t)$ with any unconditional diffusion model sampling methods like DDIM~\cite{song2021denoising} or DPM-solver~\cite{lu2022dpm}.

\newpage
\section{Experimental Details}
\label{app_expe_detail}

In this section, we will introduce more information about our experimental details. In Sec.~\ref{app_exp_domain_task}, we first introduce all the domains and tasks evaluated in our experiments. Then we briefly illustrate all the baselines compared in experiments in Sec.~\ref{app_exp_baselines}.
Then in Sec.~\ref{app_hyper_parameters}, we introduce the hyperparameters of ExDM.
Moreover, we supplement more detailed experimental results about maze2d and URLB in Sec.~\ref{app_exp_maze} and Sec.~\ref{app_exp_urlb}, respectively.
The detailed ablation studies are in Sec.~\ref{app_exp_timestep_results}. And we finally report the computing resource in Sec.~\ref{app_expe_computing_resource}.

\textbf{Codes of ExDM are provided in the Supplementary Material.}

\subsection{Domains and Tasks}
\label{app_exp_domain_task}

\paragraph{Maze2d.} 
This setting includes 7 kinds of mazes: Square-a, Square-b, Square-c, Square-d, Square-tree, Square-bottleneck, and Square-large. These mazes are two-dimensional, and agents need to explore as many states as possible during the unsupervised pre-training stage.

\begin{figure}[h]
\centering
\includegraphics[width=\linewidth]{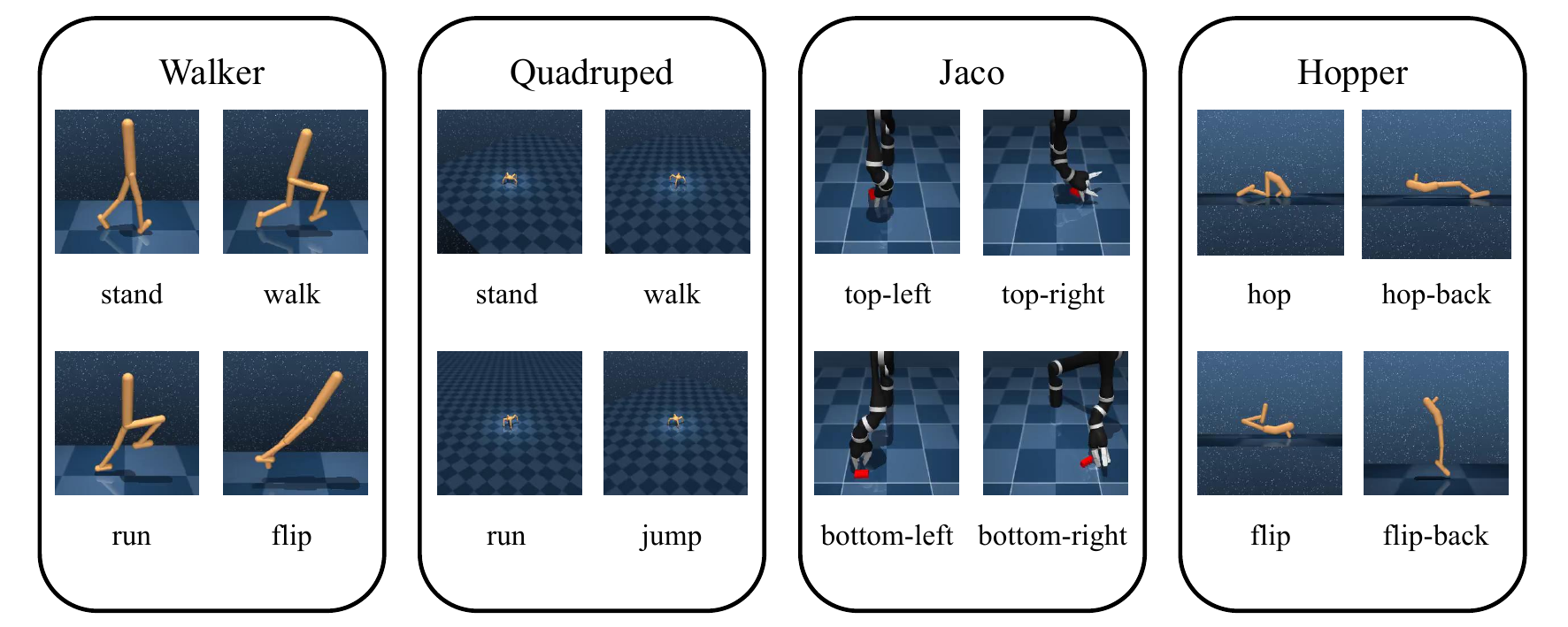}
\caption{
Illustration of domains with their downstream tasks in URLB~\cite{laskin2021urlb}. We consider 4 domains, and each domain has four downstream tasks.
}
\label{fig_dmc_tasks}
\end{figure}

\paragraph{Continuous Control.} 
Our domains of continuous control follow URLB~\cite{laskin2021urlb}, including 4 domains: Walker, Quadruped, Jaco, and Hopper, each with 4 downstream tasks from Deepmind Control Suite (DMC)~\cite{tassa2018deepmind} (we plot each task of each domain in Fig.~\ref{fig_dmc_tasks}):
\begin{itemize}
    \item \textbf{Walker} is a two-legged robot, including 4 downstream tasks: \textbf{stand}, \textbf{walk}, \textbf{run}, and \textbf{flip}. The maximum episodic length and reward for each task is 1000.
    \item \textbf{Quadruped} is a quadruped robot within a 3D space, including 4 tasks: \textbf{stand}, \textbf{walk}, \textbf{run}, and \textbf{jump}. The maximum episodic length and reward for each task is 1000.
    \item \textbf{Jaco} is a 6-DOF robotic arm with a 3-finger gripper, including 4 tasks: \textbf{reach-top-left (tl)}, \textbf{reach-top-right (tr)}, \textbf{reach-bottom-left (bl)}, and \textbf{reach-bottom-right (br)}. The maximum episodic length and reward for each task is 250.
    \item \textbf{Hopper} is a one-legged hopper robot, including 4 tasks: \textbf{hop}, \textbf{hop-backward}, \textbf{flip}, and \textbf{flip-backward}. The maximum episodic length and reward for each task is 1000.
\end{itemize}

\subsection{Baselines and Implementations}
\label{app_exp_baselines}

We first introduce all URL baselines in our experiments.

\paragraph{ICM~\cite{pathak2017curiosity}.} Intrinsic Curiosity Module (ICM) trains a forward dynamics model and designs intrinsic rewards as the prediction error of the trained dynamics model.

\paragraph{RND~\cite{burda2018exploration}.} Random Network Distillation (RND) utilizes the error between the predicted features of a trained neural network and a fixed randomly initialized neural network as the intrinsic rewards.

\paragraph{Disagreement~\cite{pathak2019self}} The Disagreement algorithm proposes a self-supervised algorithm that trains an ensemble of dynamics models and leverages the prediction variance between multiple models to estimate state uncertainty. 

\paragraph{LBS~\cite{mazzaglia2022curiosity}.} Latent Bayesian Surprise (LBS) designs the intrinsic reward as the Bayesian surprise within a latent space, i.e., the difference between prior and posterior beliefs of system dynamics.

\paragraph{DIAYN~\cite{eysenbach2018diversity}.} Diversity is All You Need (DIAYN) proposes to learn a diverse set of skills during the unsupervised pre-training stage, by maximizing the mutual information between states and skills.

\paragraph{SMM~\cite{lee2019efficient}.} State Marginal Matching (SMM) aims at learning a policy, of which the state distribution matches a given target state distribution.

\paragraph{LSD~\cite{park2022lipschitz}.} Lipschitz-constrained Skill Discovery (LSD) adopts a Lipschitz-constrained state representation function for maximizing the traveled distances of states and skills.

\paragraph{CIC~\cite{laskin2022unsupervised}.} Contrastive Intrinsic Control (CIC) leverages contrastive learning between state and skill representations, which can both learn the state representation and encourage behavioral diversity.

\paragraph{BeCL~\cite{yang2023behavior}.} Behavior Contrastive Learning (BeCL) defines intrinsic rewards as the mutual information (MI) between states sampled from the same skill, utilizing contrastive learning among behaviors.

\paragraph{CeSD~\cite{bai2024constrained}.} Constrained Ensemble exploration for Skill Discovery (CeSD) utilizes an ensemble of value functions for distinguishing different skills and encourages the agent to explore the state space with a partition based on the designed prototype.

In experiments of URLB, most baselines (ICM, RND, Disagreement, DIAYN, SMM) combined with RL backbone DDPG are directly following the official implementation in urlb~(\url{https://github.com/rll-research/url_benchmark}). For LBS, we refer to the official implementation (\url{https://github.com/mazpie/mastering-urlb}) and combine it with the codebase of urlb. For CIC, BeCL, and CeSD, we also follow their official implementations (\url{https://github.com/rll-research/cic}, \url{https://github.com/Rooshy-yang/BeCL}, \url{https://github.com/Baichenjia/CeSD}), respectively.

Below, we will list the diffusion fine-tuning baselines in our experiments.

\paragraph{DQL~\cite{wang2023diffusion}.} 
Diffusion Q-Learning (DQL) inherited the idea of policy gradient and proposes to directly backpropagate the gradient of the Q function within the actions (calculated with the diffusion action by multi-step denoising).

\paragraph{QSM~\cite{psenka2023learning}.} 
Q-Score Matching (QSM) proposes to align the score of the diffusion policy with the gradient of the learned Q function.

\paragraph{DIPO~\cite{yang2023policy}.}
Diffusion Policy for Model-free Online RL (DIPO) utilizes the Q function to optimize the actions, i.e., finding the better action with gradient ascent of the Q function, and then trains the diffusion policy to fit the ``optimized'' actions.

\paragraph{IDQL~\cite{hansen2023idql}.}
Implicit Diffusion Q-learning (IDQL) considers to sample multiple actions from the diffusion policy and then select the optimal action with the learned Q function.

We implement these methods based on their official codebases: DQL (\url{https://github.com/Zhendong-Wang/Diffusion-Policies-for-Offline-RL}), QSM (\url{https://github.com/Alescontrela/score_matching_rl}), DIPO (\url{https://github.com/BellmanTimeHut/DIPO}), and IDQL (\url{https://github.com/philippe-eecs/IDQL}).

\subsection{Hyperparameters}
\label{app_hyper_parameters}

Hyperparameters of baselines are taken from their implementations (see Appendix~\ref{app_exp_baselines} above). Here we introduce ExDM's hyperparameters. 

First, for the RL backbone DDPG, our code is based on URLB (\url{https://github.com/rll-research/url_benchmark}) and inherits DDPG's hyperparameters. For the diffusion model hyperparameters, we follow CEP~\cite{lu2023contrastive}. For completeness, we list all hyperparameters in Table~\ref{table_hyper_parameter}.

\begin{table}[h]
\centering
\footnotesize
\begin{tabular}{c|ccc}
\hline
\textbf{DDPG Hyperparameter}
& Value \\ 
\hline
Replay buffer capacity 
& $10^6$\\
Action repeat 
& 1\\
Seed frames 
& 4000\\
n-step returns 
& 3\\
Mini-batch size 
& 1024\\
Seed frames 
& 4000\\
Discount $\gamma$
& 0.99\\
Optimizer 
& Adam\\
Learning rate 
& 1e-4\\
Agent update frequency
& 2\\
Critic target EMA rate $\tau_Q$
& 0.01\\
Features dim. 
& 1024\\
Hidden dim.
& 1024\\
Exploration stddev clip
& 0.3\\
Exploration stddev value
& 0.2\\
Number of pre-training frames 
& 1$\times 10^5$ for Maze2d and 2$\times 10^6$ for URLB\\
Number of fine-turning frames
& 1$\times 10^5$ for URLB\\
\hline
\textbf{ExDM Hyperparameter} & Value \\
\hline
Diffusion SDE & VP SDE \\
$\alpha_t$ of diffusion model & $\alpha_t = -\frac{\beta_1 - \beta_0}{4}t^2 - \frac{\beta_0}{2} t, \beta_0 = 0.1, \beta_1 = 20$\\
$\sigma_t$ of diffusion model & $\sigma_t = \sqrt{1 - \alpha_t^2}$ \\
Diffusion model neural network & 3 MLPResnet Blocks, hidden$\_$dim=256, the same as IDQL~\cite{hansen2023idql} \\
Optimizer & Adam\\
Learning rate & 1e-4\\
Energy guidance model & 4-layer MLP, hidden$\_$dim=256, the same as CEP~\cite{lu2023contrastive} \\
Sampling method & DPM-Solver \\
Sampling step & 15 \\
\hline
\end{tabular}
\caption{Details of hyperparameters used for Maze2d and state-based URLB.}
\label{table_hyper_parameter}
\end{table}

\subsection{Additional Experiments in Maze}
\label{app_exp_maze}

\begin{figure}[h]
\centering
\includegraphics[width=0.95\linewidth]
{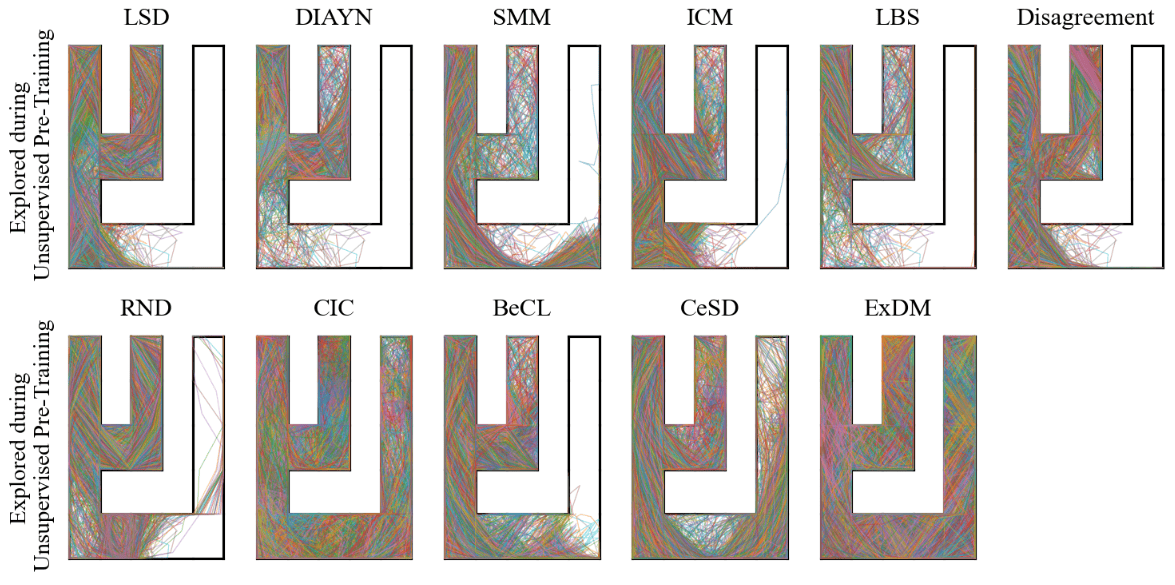}
\includegraphics[width=\linewidth]
{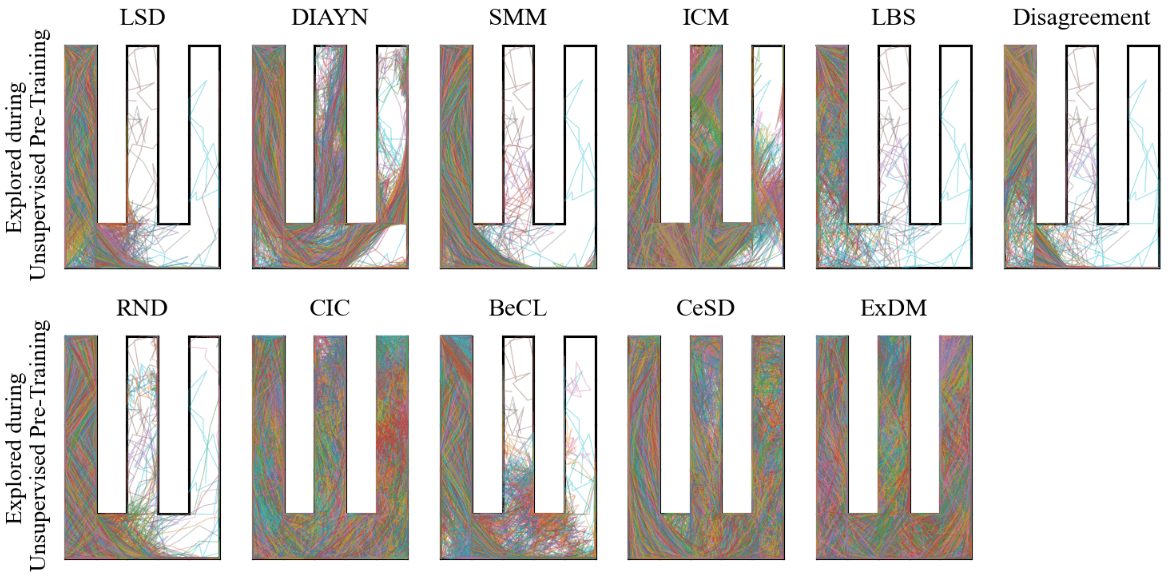}
\includegraphics[width=\linewidth]
{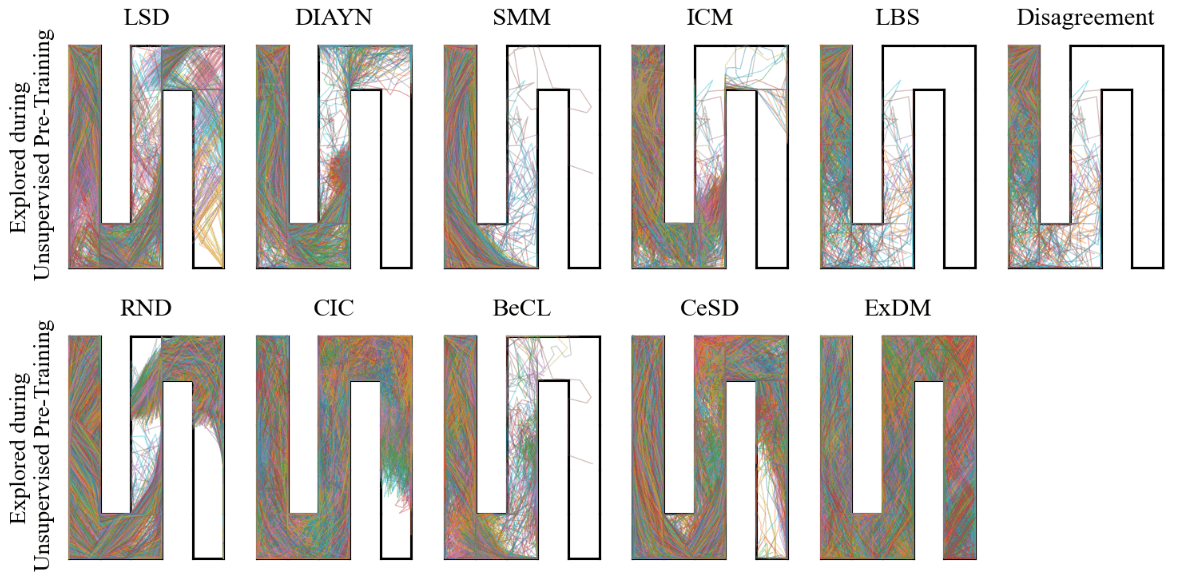}
\caption{
Visualization of explored trajectories by URL methods in \textbf{Square-a}, \textbf{Square-b}, and \textbf{Square-c} maze.
}
\label{fig_app_exp_replay_buffer_maze_a}
\end{figure}

\begin{figure}[H]
\centering
\includegraphics[width=\linewidth]
{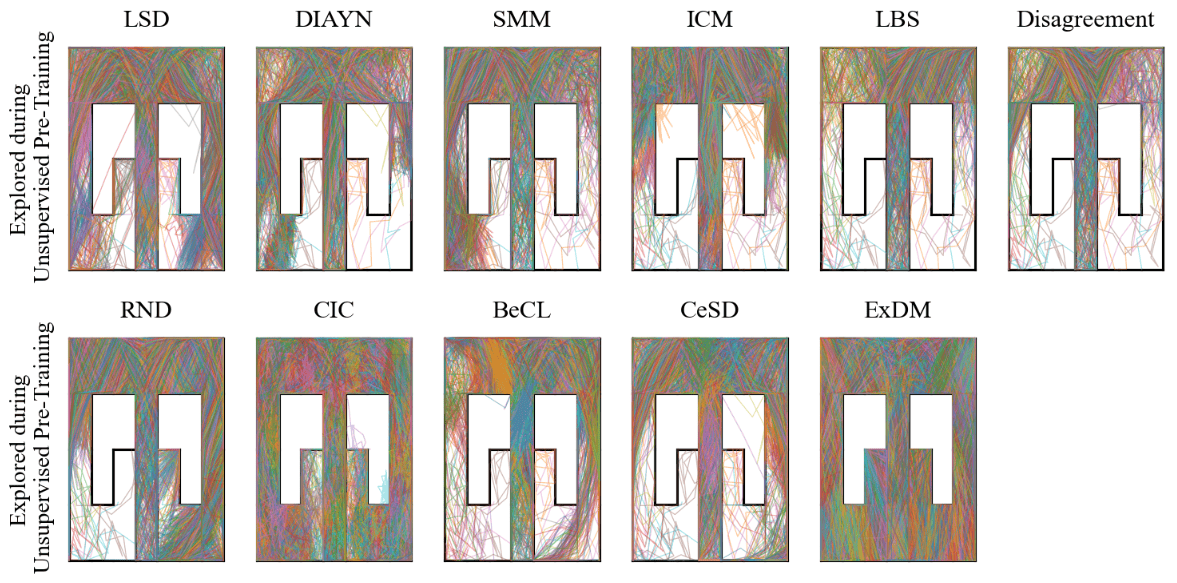}
\includegraphics[width=\linewidth]
{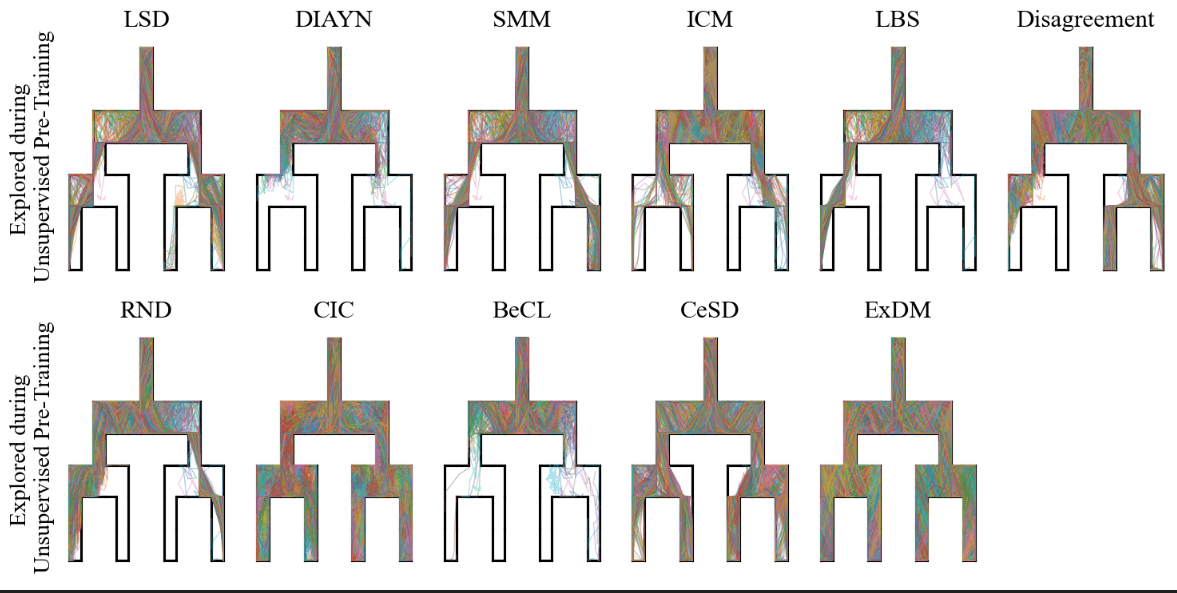}
\includegraphics[width=\linewidth]
{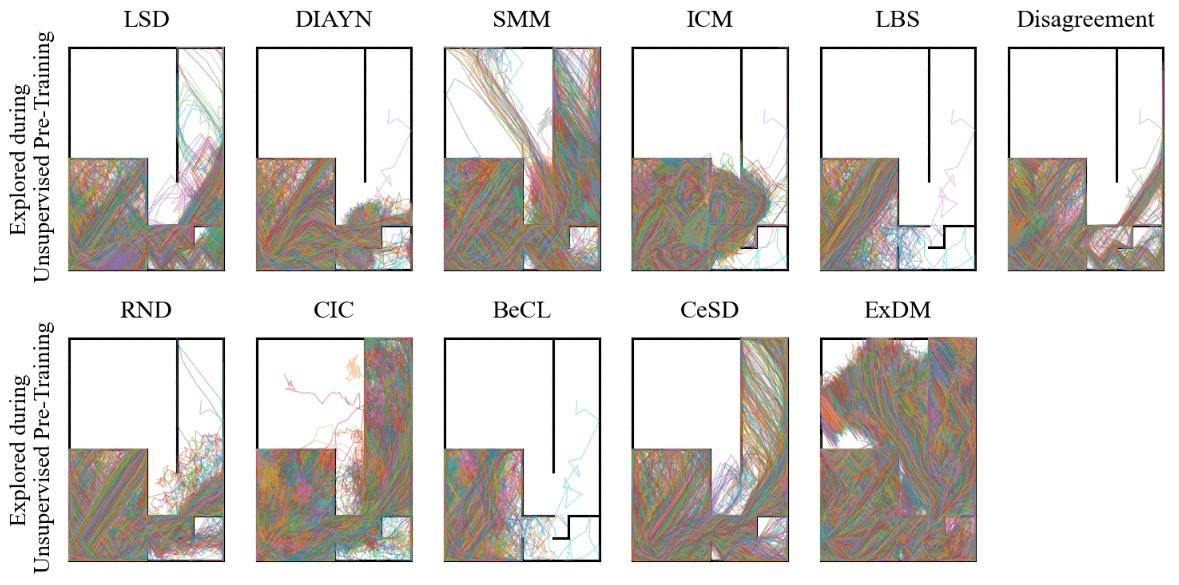}
\caption{
Visualizations of explored trajectories by URL methods in mazes \textbf{Square-d}, \textbf{Square-tree}, and \textbf{Square-bottleneck}, respectively.
}
\label{fig_app_exp_replay_buffer_maze_d}
\end{figure}

\begin{figure}[H]
\centering
\includegraphics[width=\linewidth]
{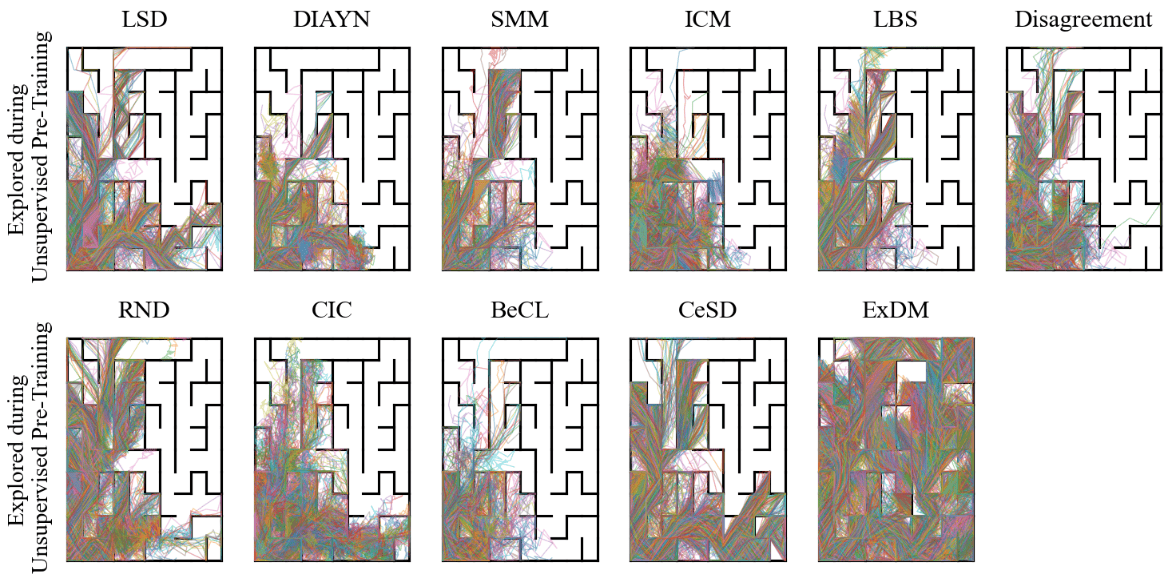}
\caption{
Visualizations of explored trajectories by URL methods in mazes \textbf{Square-large}.
}
\label{fig_app_exp_replay_buffer_maze_large}
\end{figure}

Moreover, we include the visualization of all algorithms (ICM, RND, Disagreement, LBS, DIAYN, SMM, LSD, CIC, BeCL, CeSD, and ExDM) within all 7 mazes: Square-a, Square-b, Square-c, Square-d, Square-tree, Square-bottleneck, and Square-large, in Fig.~\ref{fig_app_exp_replay_buffer_maze_a} - Fig.~\ref{fig_app_exp_replay_buffer_maze_large}, respectively. 

As shown in these figures, although baselines can explore unseen states and try to cover as many states as they can, the behaviors of baselines can not fully cover the explored replay buffer due to their limited expressive ability. 
Using the strong modeling ability of diffusion models, the pre-trained policies of ExDM can perform diverse behaviors, setting a great initialization for handling downstream tasks.

\clearpage

\clearpage

\subsection{Additional Experiments in URLB}
\label{app_exp_urlb}

In Table~\ref{table_dmc_state}, we report the detailed results of all methods in the 4 downstream tasks of 4 domains in URLB. In both the Quadruped and Jaco domains, ExDM obtains state-of-the-art performance in downstream tasks. Overall, there are the most number of downstream tasks that ExDM performs the best, and ExDM significantly outperforms existing exploration algorithms.

\begin{table}[h]
\centering
\tiny
\resizebox{1.0\textwidth}{!}{
\begin{tabular}{c|cccc|cccc|cccc|cccccc}
\toprule
Domains & \multicolumn{4}{c}{Walker} & \multicolumn{4}{c}{Quadruped} & \multicolumn{4}{c}{Jaco} & \multicolumn{4}{c}{Hopper}\\
Tasks & stand & walk & run & flip & stand & walk & run & jump  & tl & tr & bl & br & hop & hop-back & flip & flip-back\\  
\midrule
ICM
& 828.5 & 628.8 & 223.8 & 400.3
& 298.9 & 129.9 & 92.1 & 148.8
& 96.5 & 91.7 & 84.3 & 83.4
& 82.1 & 160.5 & 106.9 & 107.6\\
RND
& 878.3 & 745.4 & 348.0 & 454.1
& 792.0 & 544.5 & 447.2 & 612.0
& 98.7 & 110.3 & 107.0 & 105.2
& 83.3 & \textbf{267.2} & \textbf{132.5} & 184.0\\
Disagreement
& 749.5 & 521.9 & 210.5 & 340.1
& 560.8 & 382.3 & 361.9 & 427.9
& 142.5 & 135.1 & 129.6 & 118.1
& 86.2 & 255.6 & 113.0 & \textbf{215.3}\\
LBS
& 594.9 & 603.2 & 138.8 & 375.3
& 413.0 & 253.2 & 203.8 & 366.6
& 166.5 & 153.8 & 129.6 & 139.6
& 24.8 & 240.2 & 88.9 & 105.6\\
\hline
DIAYN
& 721.7 & 488.3 & 186.9 & 317.0
& 640.8 & 525.1 & 275.1 & 567.8
& 29.7 & 15.6 & 30.4 & 38.6
& 1.7 & 10.8 & 0.7 & 0.5\\
SMM
& 914.3 & 709.6 & 347.4 & 442.7
& 223.9 & 93.8 & 91.6 & 96.2
& 57.8 & 30.1 & 34.8 & 45.0
& 29.3 & 61.4 & 47.0 & 29.7\\
LSD
& 770.2 & 532.3 & 167.1 & 309.7
& 319.4 & 186.3 & 179.6 & 283.5
& 11.6 & 33.6 & 22.5 & 6.7
& 12.0 & 6.6 & 2.9 & 12.2 \\
CIC
& \textbf{941.1} & \textbf{883.1} & \textbf{399.0} & \textbf{687.2}
& 789.1 & 587.8 & 475.1 & 630.6
& 148.8 & 167.6 & 122.3 & 145.9
& 82.7 & 191.6 & 96.2 & 161.3\\
BeCL
& \textbf{951.7} & \textbf{912.7} & \textbf{408.6} & 626.2
& 798.7 & 694.9 & 391.7 & 645.5
& 114.2 & 132.2 & 117.7 & 144.7
& 37.1 & 68.3 & 73.6 & 142.7\\
CeSD
& 884.0 & 838.7 & 325.2 & 570.9
& \textbf{886.5} & 763.4 & \textbf{636.4} & \textbf{759.1} 
& 155.7 & 170.2 & 137.3 & 117.9 
& \textbf{118.4} & 155.7 & 46.4 & 183.6 \\
\hline
ExDM (Ours)
& \textbf{905.9} & \textbf{874.1} & \textbf{389.5} & 572.1
& \textbf{915.4} & \textbf{873.6} & 569.5 & \textbf{755.2}
& \textbf{173.5} & \textbf{202.6} & \textbf{170.3} & \textbf{170.5}
& \textbf{107.6} & \textbf{275.8} & \textbf{155.3} & \textbf{220.1}\\
\bottomrule
\end{tabular}
}
\caption{\textbf{Detailed results in URLB of different pre-trained methods that fine-tune Gaussian policies with DDPG}. Average cumulative reward (mean of 10 seeds) of the best policy.}
\label{table_dmc_state}
\end{table}

\begin{table}[h]
\centering
\footnotesize
\begin{tabular}{c|cccc}
\toprule
Metrics & Median & IQM & Mean & Optimality Gap\\  
\midrule
ICM
& 0.40 & 0.38 & 0.39 & 0.61 \\
RND
& 0.59 & 0.59 & 0.59 & 0.41 \\
Disagreement
& 0.54 & 0.53 & 0.57 & 0.46 \\
LBS
& 0.48 & 0.45 & 0.47 & 0.53 \\
\hline
DIAYN
& 0.29 & 0.24 & 0.29 & 0.71 \\
SMM
& 0.29 & 0.21 & 0.29 & 0.71 \\
LSD
& 0.21 & 0.14 & 0.21 & 0.79 \\
CIC
& 0.66 & 0.68 & 0.66 & 0.35 \\
BeCL
& 0.59 & 0.62 & 0.60 & 0.40\\
CeSD
& 0.67 & 0.71 & 0.67 & 0.33\\
\hline
ExDM (Ours)
& \textbf{0.78} & \textbf{0.80} & \textbf{0.77} & \textbf{0.23} \\
\bottomrule
\end{tabular}
\caption{\textbf{Aggregate metrics~\cite{agarwal2021deep} in URLB}. For every algorithm, there are 4 domains, each trained with 10 seeds and fine-tuned under 4 downstream tasks, thus each statistic for every method has 160 runs.}
\label{table_dmc_state_rliable}
\end{table}

Moreover, we further provide the detailed metrics of all methods within URLB in Table~\ref{table_dmc_state_rliable}. As shown here, ExDM significantly outperforms all baselines, for example, ExDM' IQM is larger than the second-best method, CeSD, by 13$\%$.


\subsection{Ablation of timesteps in URLB}
\label{app_exp_timestep_results}

\begin{figure}[h]
    \centering
    \begin{minipage}{0.48\textwidth}
     \includegraphics[height=4.2cm,width=6.0cm]{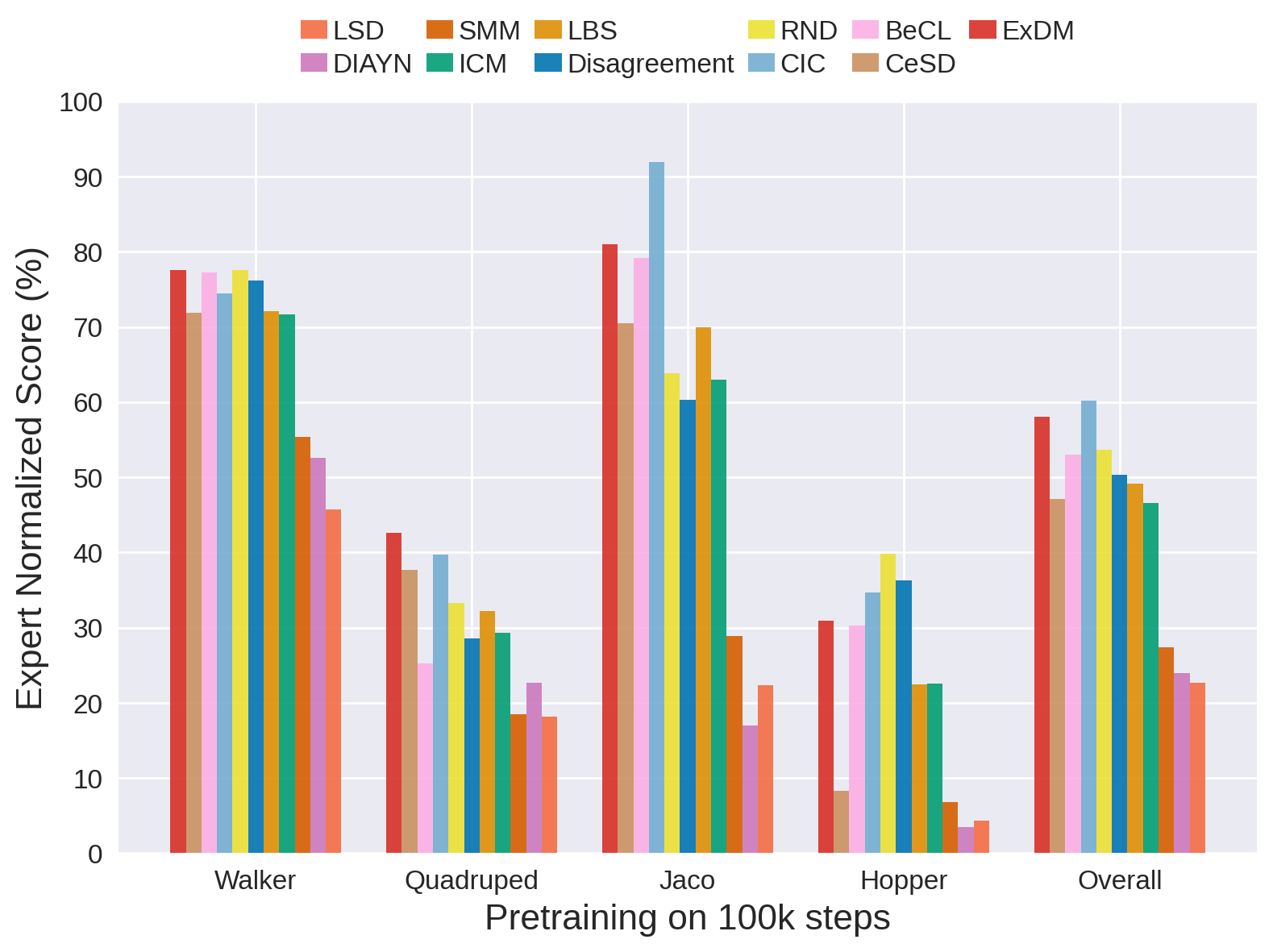}
    \end{minipage}
     \begin{minipage}{0.48\textwidth}
     \includegraphics[height=4.2cm,width=6.0cm]{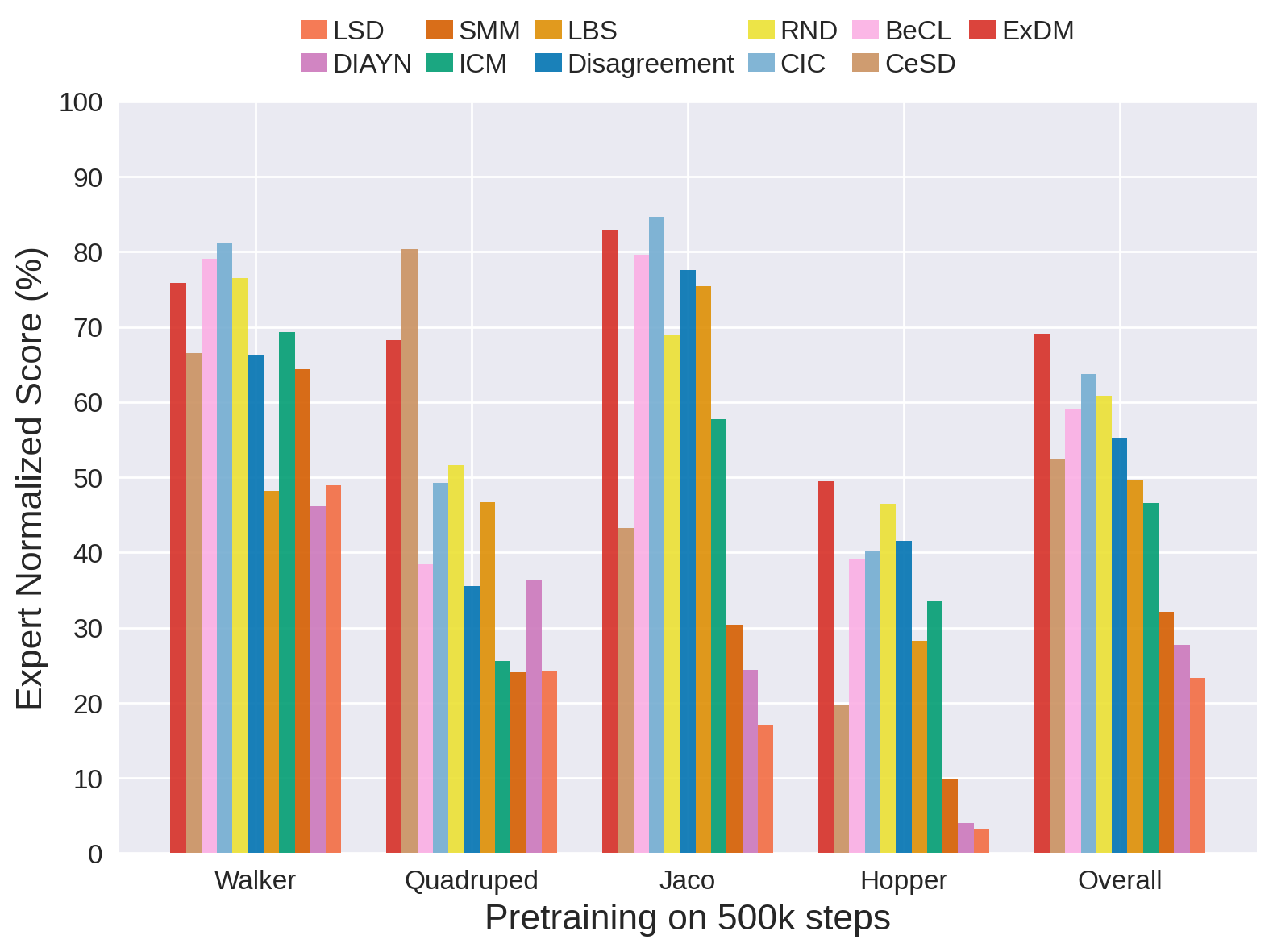}
    \end{minipage}
    \begin{minipage}{0.48\textwidth}
     \includegraphics[height=4.2cm,width=6.0cm]{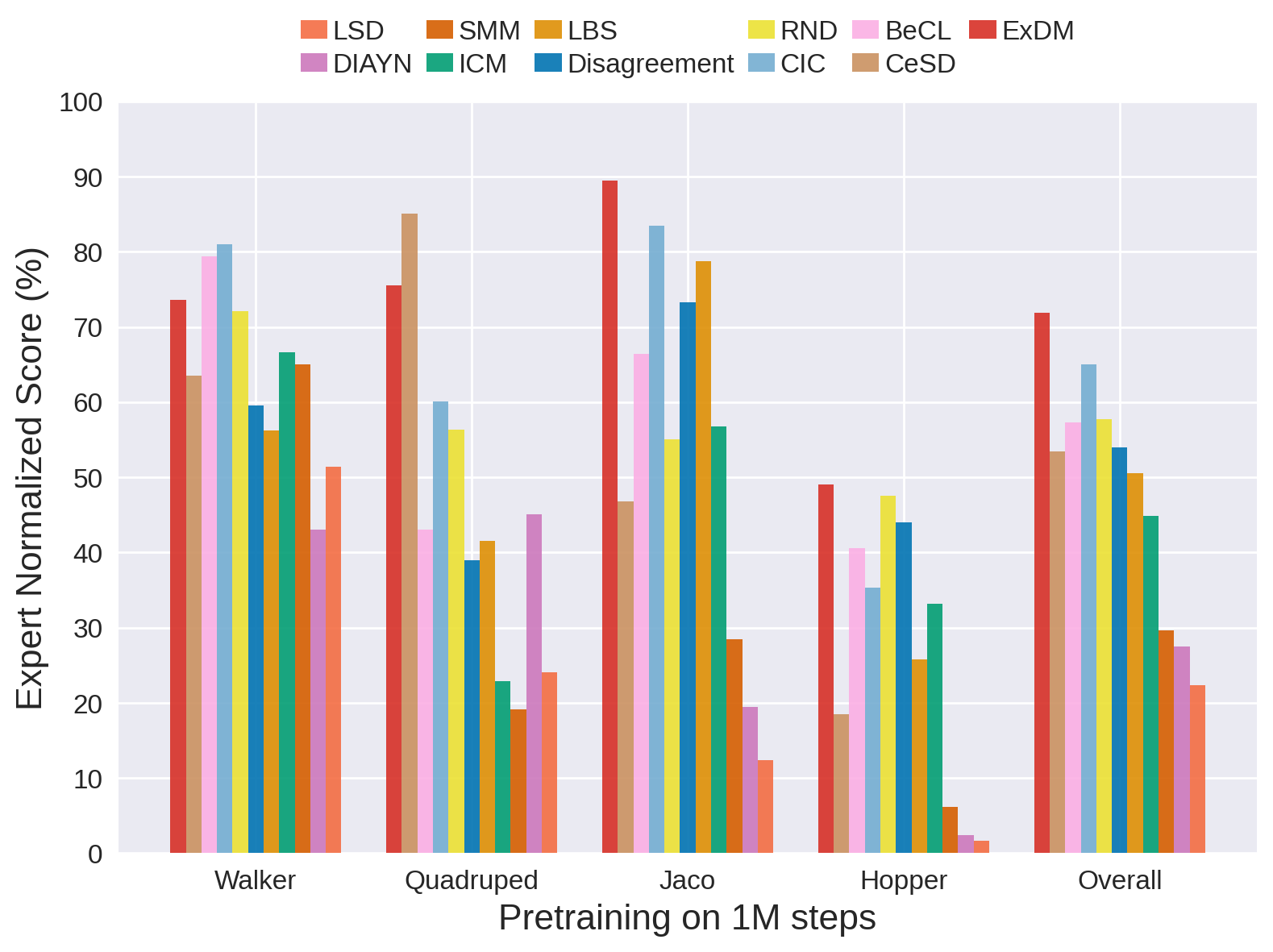}
    \end{minipage}
     \begin{minipage}{0.48\textwidth}
     \includegraphics[height=4.2cm,width=6.0cm]{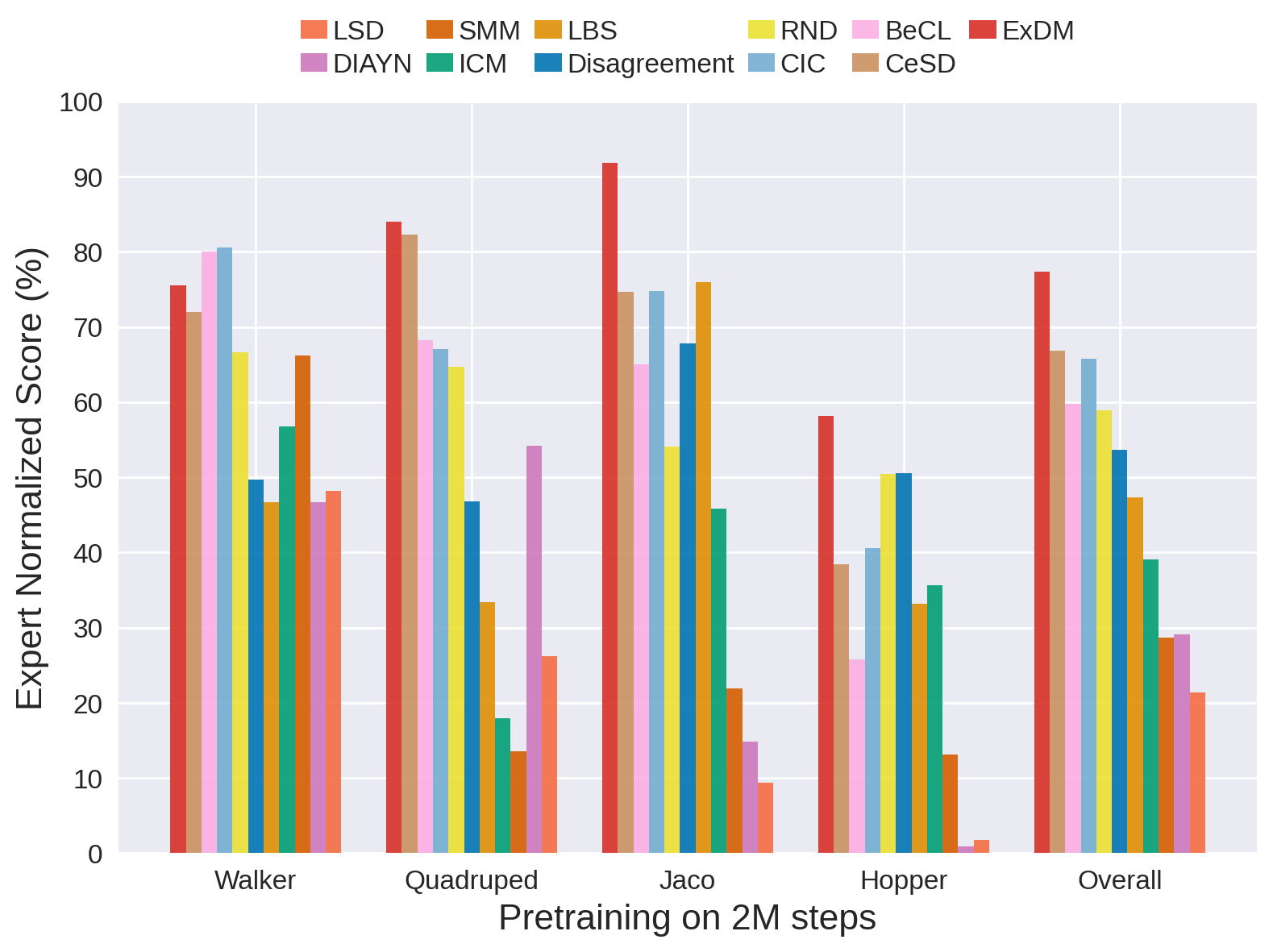}
    \end{minipage}
    \caption{\textbf{Ablation study of pre-training steps in URLB}.}
    \label{fig_timestep_all}
\end{figure}

In Figure~\ref{fig_timestep_all}, we show additional results about the performance in four domains of URLB for different algorithms and pre-training timesteps. Overall, ExDM outperforms all methods, while CIC and CeSD are still competitive on some domains. 

\subsection{Computing Resource}
\label{app_expe_computing_resource}

In experiments, all agents are trained by GeForce RTX 2080 Ti with Intel(R) Xeon(R) Silver 4210 CPU @ 2.20GHz. 
In maze2d / urlb, pre-training ExDM (each seed, domain) takes around 0.5 / 2 days, respectively.

\section{Broader Impact}
\label{sec_broader_impact}

Designing generalizable agents for varying tasks is one of the major concerns in reinforcement learning. This work focuses on utilizing diffusion policies for exploration and proposes a novel algorithm ExDM. One of the potential negative impacts is that algorithms mainly use deep neural networks, which lack interoperability and may face robustness issues. There are no serious ethical issues as this is basic research.

\end{document}